\documentclass[journal]{IEEEtran}
\usepackage{hyperref}
\usepackage{url}

\usepackage{subfigure}
\usepackage{epsfig}
\usepackage{epstopdf}
\usepackage{graphicx}
\usepackage{amssymb}
\usepackage{amsmath}
\usepackage{amsthm}
\usepackage{color}
\usepackage{booktabs}
\usepackage{array}

\DeclareMathOperator*{\mytr}{tr}
\DeclareMathOperator*{\rank}{rank}

\DeclareMathOperator*{\myvec}{vec}
\DeclareMathOperator*{\mydiag}{diag}
\DeclareMathOperator*{\spark}{spark}
\DeclareMathOperator*{\myspan}{span}
\usepackage{enumitem}
\newcommand{\myendofproof}[0]{\hfill $\blacksquare$ \newline}

\input{def.set}

\ifCLASSINFOpdf
\else
\fi

\begin{document}
\title{Exploring Algorithmic Limits of  Matrix Rank Minimization under Affine Constraints}

\author{Bo~Xin
        and~David~Wipf% <-this % stops a space
\thanks{B. Xin is with the Department
of Electrical Engineering and Computer Science, Peking University, Beijing, China. e-mail: jimxinbo@gmail.com}% <-this % stops a space
\thanks{D. Wipf is with the Visual Computing group, Microsoft Research, Beijing, China. e-mail: davidwip@microsoft.com}}% <-this % stops a space
\maketitle

\begin{abstract}
Many applications require recovering a matrix of minimal rank within an affine constraint set, with matrix completion a notable special case.  Because the problem is NP-hard in general, it is common to replace the matrix rank with the nuclear norm, which acts as a convenient convex surrogate. While elegant theoretical conditions elucidate when this replacement is likely to be successful, they are highly restrictive and convex algorithms fail when the ambient rank is too high or when the constraint set is poorly structured.  Non-convex alternatives fare somewhat better when carefully tuned; however, convergence to locally optimal solutions remains a continuing source of failure.  Against this backdrop we derive a deceptively simple and parameter-free probabilistic PCA-like algorithm that is capable, over a wide battery of empirical tests, of successful recovery even at the theoretical limit where the number of measurements equal the degrees of freedom in the unknown low-rank matrix.  Somewhat surprisingly, this is possible even when the affine constraint set is highly ill-conditioned.  While proving general recovery guarantees remains evasive for non-convex algorithms, Bayesian-inspired or otherwise, we nonetheless show conditions whereby the underlying cost function has a unique stationary point located at the global optimum; no existing cost function we are aware of satisfies this same property.  We conclude with a simple computer vision application involving image rectification and a standard collaborative filtering benchmark.
\end{abstract}


% \fi
\IEEEpeerreviewmaketitle

%============================

\section{Introduction}
\label{sec:intro}

Recently there has been a surge of interest in finding minimum rank matrices subject to some problem-specific constraints often characterized as an affine set \cite{candes2009exact,hu2012fast,liu2013robust,zhang2012tilt,lu2014generalized,mohan2012iterative}.  Mathematically this involves solving
\begin{equation} \label{eq:affine_rank_problem}
\min_{\bX} ~~ \rank[\bX] \hspace*{0.5cm} \mbox{s.t. } \bb = \mathcal{A}(\bX),
\end{equation}
where $\bX \in \mathbb{R}^{n\times m}$ is the unknown matrix, $\bb \in \mathbb{R}^p$ represents a vector of observations and $\calA : \mathbb{R}^{n\times m} \rightarrow \mathbb{R}^p$ denotes a linear mapping.  An important special case of (\ref{eq:affine_rank_problem}) commonly applied to collaborative filtering is the matrix completion problem
\begin{equation} \label{eq:mc}
    \min_{\bX} ~~  {\rank[\bX]} \hspace*{0.5cm} \mbox{s.t.}~~ \bX_{ij} = (\bX_0)_{ij},~~(i,j) \in \Omega,
\end{equation}
where $\bX_0$ is a low-rank matrix we would like to recover, but we are only able to observe elements from the set $\Omega$ \cite{candes2009exact,hu2012fast}.  Unfortunately however, both this special case and the general problem (\ref{eq:affine_rank_problem}) are well-known to be NP-hard, and the rank penalty itself is non-smooth.  Consequently, a popular alternative is to instead compute
\begin{equation} \label{eq:general_low_rank}
\min_{\bX} ~~ \sum_{i} f(\sigma_i[\bX]) \hspace*{0.5cm} \mbox{s.t. } \bb = \mathcal{A}(\bX),
\end{equation}
where $\sigma_i[\bX]$ denotes the $i$-th singular value of $\bX$ and $f$ is usually a concave, non-decreasing function (or nearly so).  In the special case where $f(z) = I[z \neq 0]$ (i.e., an indicator function) we retrieve the matrix rank; however, smoother surrogates such as $f(z) = \log z$ or $f(z) = z^q$ with  $q\leq 1$ are generally preferred for optimization purposes.  When $f(z) = z$, (\ref{eq:general_low_rank}) reduces to convex nuclear norm minimization.  A variety of celebrated theoretical results have quantified specific conditions, heavily dependent on the singular values of matrices in the nullspace of $\calA$,  where the minimum nuclear norm solution is guaranteed to coincide with that of minimal rank \cite{candes2009exact,liu2013robust,mohan2012iterative}.  However, these guarantees typically only apply to a highly restrictive set of rank minimization problems, and in a practical setting non-convex algorithms can succeed in a much broader range of conditions \cite{hu2012fast,lu2014generalized,mohan2012iterative}.

In Section \ref{sec:related_work} we will summarize state-of-the-art non-convex rank minimization algorithms that operate under affine constraints and point out some of their shortcomings.  This will be followed in Section \ref{sec:alg} by the derivation of an alternative approach using Bayesian modeling techniques adapted from probabilistic PCA \cite{Tipping1999PPCA}.  Section \ref{sec:analysis} will then describe connections with nuclear norm minimization, convergence issues, and properties of global and local solutions.  The latter includes special cases whereby any stationary point of the intrinsic cost function is guaranteed to have optimal rank, illustrating an underlying smoothing mechanism which leads to success over competing methods.  We next discuss algorithmic enhancements in Section \ref{sec:symm} that further improve recovery performance in practice.  Section \ref{sec:expr} contains a wide variety of numerical comparisons that highlight the efficacy of this algorithm, while Section \ref{sec:app} presents a computer vision application involving image rectification and a standard collaborative filtering benchmark.  Technical proofs and algorithm update rule details are contained in the Appendix.

%In the supplemental file we provide technical proofs  and illustrations, as well as a computer vision application involving image rectification and a standard collaborative filtering benchmark.  Before proceeding, we summarize two main contributions as follows:

Before proceeding, we highlight several main contributions as follows:

\begin{itemize}
\item Bayesian inspiration can take uncountably many different forms and parameterizations, but the devil is in the details and existing methods offer little opportunity for both theoretical inquiry and substantial performance gains solving (\ref{eq:affine_rank_problem}). In this regard, we apply carefully-tailored modifications to a veteran probabilistic PCA model leading to systematic analytical and empirical insights and advantages.  Model justification is ultimately based on such meticulous technical considerations rather than merely the presumed qualitative legitimacy of any underlying prior distributions.

\item Non-convex algorithms have demonstrated some improvement in estimation accuracy over the celebrated convex nuclear norm; however, this typically requires the inclusion of one or more additional tuning parameters to incrementally inject additional objective function curvature and avoid bad local solutions.  In contrast, for solving (\ref{eq:affine_rank_problem}) our non-convex Bayesian-inspired algorithm requires no such parameters at all, and noisy relaxations necessitate only a single, standard trade-off parameter balancing data-fit and minimal rank.\footnote{While not our emphasis here, similar to other Bayesian frameworks, even this trade-off parameter can ultimately be learned from the data if a true, parameter-free implementation is desired across noise levels.}

\item Over a wide battery of controlled experiments with ground-truth data, our approach outperforms all existing algorithms that we are aware of, Bayesian, non-convex, or otherwise. This includes direct head-to-head comparisons using the exact experimental designs and code prepared by original authors with carefully tuned parameters.  In fact, even when $\mathcal{A}$ is ill-conditioned we are consistently able to solve (\ref{eq:affine_rank_problem}) right up to the theoretical limit of any possible algorithm, which has never been demonstrated previously.
\end{itemize}

\section{Related Work}
\label{sec:related_work}

Here we focus on a few of the latest and most effective rank minimization algorithms, all developed within the last few years and evaluated favorably against the state-of-the-art.

\subsection{General Non-Convex Methods}

%, which coincides with the limit as $q \rightarrow 0$ (up to an inconsequential scaling and translation),

In the non-convex regime, effective optimization strategies attempt to at least locally minimize (\ref{eq:general_low_rank}), often exceeding the performance of the convex nuclear norm. For example, \cite{mohan2012iterative} derives a family of \emph{iterative reweighted least squares} (IRLS) algorithms applied to $f(z) = (z^2 + \gamma)^{q/2}$ with $q,\gamma > 0$ as tuning parameters.  A related penalty also considered is $f(z) = \log(z^2 + \gamma)$, which maintains an intimate connection with rank given that
\begin{equation}
\log z = \lim_{q \rightarrow 0} q^{-1}( z^q - 1) ~~\mbox{and}~~ \lim_{q \rightarrow 0} z^q = I[z \neq 0],
\end{equation}
where $I$ is a standard indicator function.  Consequently, when $\gamma$ is small,  $\sum_i \log(\sigma_i[\bX]^2 + \gamma)$ behaves much like a scaled and translated version of the rank, albeit with nonzero gradients away from zero.

The IRLS0 algorithm from \cite{mohan2012iterative} represents the best-performing special case of the above, where $\sum_i \log(\sigma_i[\bX]^2 + \gamma)$ is minimized using a homotopy continuation scheme merged with IRLS.  Here a fixed $\gamma$ is replaced with a decreasing sequence $\{\gamma^k\}$, the rationale being that when $\gamma^k$ is large, the cost function is relatively smooth and devoid of local minima.  As the iterations $k$ progress, $\gamma^k$ is reduced, and the cost behaves more like the matrix rank function.  However, because now we are more likely to be within a reasonably good basin of attraction, spurious local minima are more easily avoided.  The downside of this procedure is that it requires a pre-defined heuristic for reducing $\gamma^k$, and this schedule may be problem specific.  Moreover, there is no guarantee that a global solution will ever be found.

In a related vein, \cite{lu2014generalized} derives a family of \emph{iterative reweighted nuclear norm} (IRNN) algorithms that can be applied to virtually any concave non-decreasing function $f$, even when $f$ is non-smooth, unlike IRLS.  For effective performance however the authors suggest a continuation strategy similar to IRLS0.  Moreover, additional tuning parameters are required for different classes of functions $f$ and it remains unclear which choices are optimal.  While the reported results are substantially better than when using the convex nuclear norm, in our experiments IRLS0 seems to perform slightly better, possibly because the quadratic least squares inner loop is less aggressive in the initial stages of optimization than weighted nuclear norm minimization, leading to a better overall trajectory.  Regardless, all of these affine rank minimization algorithms fail well before the theoretical recovery limit is reached, when the number of observations $p$ equals the number of degrees of freedom in the low-rank matrix we wish to recover.  Specifically, for an $n\times m$, rank $r$ matrix, the number of degrees of freedom is given by $r(m+n) - r^2$, hence $p = r(m+n) - r^2$ is the best-case boundary.  In practice if $\calA$ is ill-conditioned or degenerate the achievable limit may be more modest.

A third approach relies on replacing the convex nuclear norm with a truncated non-convex surrogate \cite{hu2012fast}. While some competitive results for image impainting via matrix completion are shown, in practice the proposed algorithm has many parameters to be tuned via cross-validation.  Moreover, recent comparisons contained in \cite{lu2014generalized} show that default settings perform relatively poorly.

Finally, a somewhat different class of non-convex algorithms can be derived using a straightforward application of alternating minimization \cite{jain2013low}.  The basic idea is to assume $\bX = \bU \bV^T$ for some low-rank matrices $\bU$ and $\bV$ and then solve
\begin{equation}
\min_{\bU,\bV} \|b - \mathcal{A}(\bU \bV^T) \|_{\mathcal{F}}
\end{equation}
via coordinate decent.  The downside of this approach is that it requires that $\bU$ and $\bV$ be parameterized with the correct rank.  In contrast, our emphasis here is on algorithms that require no prior knowledge whatsoever regarding the true rank. Regardless, experimental results suggest that even when the correct rank is provided, these algorithms still cannot match the performance of our proposal.  Moreover, from a generalization standpoint, these rank-aware variant are not suitable for embedding in a larger system with multiple low-rank components to estimate, since it is typically not feasible to simultaneously tune multiple rank parameters.  In contrast, our method can be naturally extended for this purpose.

\subsection{Bayesian Methods}

From a probabilistic perspective, previous work has applied Bayesian formalisms to rank minimization problems, although not specifically within an affine constraint set.  For example, \cite{babacan2012sparse,ding2011bayesian,wipf2012non} derive robust PCA algorithms built upon the linear summation of a rank penalty and an element-wise sparsity penalty.  In particular, \cite{ding2011bayesian} applies an MCMC sampling approach for posterior inference, but the resulting iterations are not scalable, subjectable to detailed analysis, nor readily adaptable to affine constraints.  In contrast, \cite{babacan2012sparse} applies a similar probabilistic model but performs inference using a variational mean-field approximation.  While the special case of matrix completion is considered, from an empirical standpoint its estimation accuracy is not competitive with the state-of-the-art non-convex algorithms mentioned above.  Finally, without the element-wise sparsity component intrinsic to robust PCA (which is not our focus here), \cite{wipf2012non} simply collapses to a regular PCA model with a closed-form solution, so the challenges faced in solving (\ref{eq:affine_rank_problem}) do not apply.  Consequently, general affine constraints really are a key differentiating factor.

%In general though, none of these Bayesian-inspired algorithms have been augmented and rigorously analyzed or tested in the context of rank minimization with general affine constraints.  Moreover, the limited analysis that does exist in \cite{wipf2012non} actually just follows from the element-wise sparsity component intrinsic to robust PCA, without which the model effectively reduces to regular PCA devoid of any theoretical uncertainty.  So the general affine constraints really are the key differentiating factor.

From a motivational angle, the basic probabilistic model with which we begin our development can be interpreted as a carefully re-parameterized generalization of the probabilistic PCA model from \cite{Tipping1999PPCA}.  This will ultimately lead to a non-convex algorithm devoid of the heuristic tuning strategies mentioned above, but nonetheless still uniformly superior in terms of estimation accuracy.  We emphasize that, although we employ a Bayesian entry point for our algorithmic strategy, final justification of the underlying model will be entirely based on properties of the underlying cost function that emerges, rather than any putative belief in the actual validity of the assumed prior distributions or likelihood function.  This is quite unlike the vast majority of existing Bayesian approaches.

%, although the later is not motivated in the context of rank minimization.

%We would argue that one of our most significant contributions herein is to rigorously demonstrate, both theoretically and empirically, that such a simple origin can lead to state-of-the-art recovery results when adapted judiciously.

% We will discuss this connection more in Section ???.

\subsection{Analytical Considerations}
Turning to analytical issues, a number of celebrated theoretical results dictate conditions whereby substitution of the rank function with the convex nuclear norm in (\ref{eq:affine_rank_problem}) is nonetheless guaranteed to still produce the minimal rank solution.  For example, if $\mathcal{A}$ is a Gaussian iid measurement ensemble and $\bX_0 \in \mathbb{R}^{n\times n}$ represents the optimal solution to (\ref{eq:affine_rank_problem}) with $\mbox{rank}[\bX_0] = r$, then with high probability as the problem dimensions grow large, the minimum nuclear norm feasible solution will equal $\bX_0$ if the number of measurements $p$ satisfies $p \geq 3 r(2n - r)$ \cite{chandrasekaran2012convex}.

The limitation of this type of result is two-fold.  First, in the above situation the true minimum rank solution only actually requires $p \geq r(2n-r)$ measurements to be recoverable via brute force solution of (\ref{eq:affine_rank_problem}), and the remaining difference of a factor of three can certainly be considerable in many practical situations (e.g., requiring 300 measurements is far more laborious than only needing 100 measurements).  Secondly though, and far more importantly, all existing provable recovery guarantees place extremely strong restrictions on the structure of $\mathcal{A}$, e.g., strong restrictions on the singular value decay of matrices in the nullspace of $\mathcal{A}$.  Such conditions are unlikely to ever hold in realistic application settings, including the image rectification example we describe in Section \ref{sec:image_rectification} (in fact, these conditions are usually incapable of even being checked).  In contrast, the algorithm we propose is empirically observed to only require the theoretically minimal number of measurements even when such nullspace conditions are violated in many cases.  While a general theoretical guarantee of this sort is obviously not possible, we do nonetheless provide several supporting theoretical results indicative of why such performance is at least empirically obtainable.

\section{Alternative Algorithm Derivation}
\label{sec:alg}

In this section we first detail our basic distributional assumptions followed by development of the associated update rules for inference.

\subsection{Basic Model}

In contrast to the majority of existing algorithms organized around practical solutions to (\ref{eq:general_low_rank}), here we adopt an alternative, probabilistic starting point.  We first define the Gaussian likelihood function
\begin{equation}
p(\bb|\bX;\calA,\lambda) \hspace*{0.2cm} \propto \hspace*{0.2cm} \exp\left[ -\frac{1}{2\lambda} \Vert \calA(\bX)-\bb \Vert_2^2 \right],
\end{equation}
noting that in the limit as $\lambda \rightarrow 0$ this will enforce the same constraint set as in (\ref{eq:affine_rank_problem}).  Next we define an independent, zero-mean Gaussian prior distribution with covariance $\nu_i \bPsi$ on each column of $\bX$, denoted $\bx_{:i}$ for all $i = 1,\ldots, m$.  This produces the aggregate prior on $\bX$ given by
\begin{equation} \label{eq:basic_prior}
p(\bX;\bPsi,\bnu) \hspace*{0.2cm} = \hspace*{0.2cm}  \prod_i \calN \left(\bx_{:i};\mathbf{0},\nu_i \bPsi \right)  \hspace*{0.2cm} \propto \hspace*{0.2cm}  \exp\left[\bx^{\top} \bar{\bPsi}^{-1} \bx\right],
\end{equation}
where $\bPsi \in \mathbb{R}^{n \times n}$ is a positive semi-definite symmetric matrix,\footnote{Technically $\bPsi$ must be positive definite for the inverse in (\ref{eq:basic_prior}) to be defined.  However, we can accommodate the semi-definite case using the following convention. Without loss of generality assume that $\bar{\bPsi} = \bR \bR^{\top}$ for some matrix $\bR$.  We then qualify that $p(\bX;\bPsi,\bnu) = 0$ if $\bx \notin \mbox{span}[\bR]$, and $p(\bX;\bPsi,\bnu) \propto  \exp\left[\bx^{\top} (\bR^{\top})^{\dag} \bR^{\dag} \bx\right]$ otherwise.  Equivalently, throughout the paper for convenience (and with slight abuse of notation) we define  $\bx^{\top} \bar{\bPsi}^{-1} \bx = \infty$ when $\bx \notin \mbox{span}[\bR]$, and $\bx^{\top} \bar{\bPsi}^{-1} \bx = \bx^{\top} (\bR^{\top})^{\dag} \bR^{\dag} \bx$ otherwise.  This will come in handy, for example, when interpreting the bound in (\ref{eq:term1_upper_bound}) below.} $\bnu = [\nu_1,\ldots,\nu_m]^{\top}$ is a non-negative vector, $\bx = \myvec[\bX]$ (column-wise vectorization), and $\bar{\bPsi} = \mydiag[\bnu] \otimes \bPsi$, with $\otimes$ denoting the Kronecker product.  It is important to stress here that we do not necessarily believe that the unknown $\bX$ actually follows such a Gaussian distribution per se.  Rather, we adopt (\ref{eq:basic_prior}) primarily because it will lead to an objective function with desirable properties related to solving (\ref{eq:affine_rank_problem}).

Moving forward, given both likelihood and prior are Gaussian, the posterior $p(\bX|\bb;\bPsi,\bnu,\calA,\lambda)$ is also Gaussian, with mean given by an $\hat{\bX}$ such that
\begin{equation} \label{eq:posterior_mean}
\hat{\bx} = \myvec[\hat{\bX}] = \bar{\bPsi} \bA^{\top} \left(\lambda \bI + \bA \bar{\bPsi} \bA^{\top} \right)^{-1} \bb.
\end{equation}
Here $\bA \in \mathbb{R}^{p\times nm}$ is a matrix defining the linear operator $\calA$ such that $\bb = \bA \bx$ reproduces the feasible region in (\ref{eq:affine_rank_problem}).  From this expression it is clear that, if $\bPsi$ represents a low-rank covariance matrix, then each column of $\hat{\bX}$ will be constrained to a low-dimensional subspace resulting overall in a low-rank estimate as desired.  Of course for this simple strategy to be successful we require some way of determining a viable $\bPsi$ and the scaling vector $\bnu$.

A common Bayesian strategy in this regard is to marginalize over $\bX$ and then maximize the resulting likelihood function with respect to $\bPsi$ and $\bnu$ \cite{tipping2001sparse,wipf2012non,wipf2011latent}.  This involves solving
\begin{equation} \label{eq:marginal_like}
\max_{\bPsi \in H^+,\bnu \geq 0} \int  p(\bb|\bX;\calA,\lambda)p(\bX;\bPsi,\bnu) d \bX,
\end{equation}
where $H^+$ denotes the set of positive semi-definite and symmetric $n\times n$ matrices.  After a $-2\log$ transformation and application of a standard convolution-of-Gaussians integration, solving (\ref{eq:marginal_like}) is equivalent to minimizing the cost function
\begin{equation} \label{eq:BARM_cost}
\calL(\bPsi,\bnu) = \bb^{\top} \bSigma_b^{-1} \bb + \log \left|  \bSigma_b \right|,
\end{equation}
where
\begin{equation}
\bSigma_b = \bA \bar{\bPsi} \bA^{\top} + \lambda \bI ~~\mbox{and}~~ \bar{\bPsi} = \mydiag[\bnu] \otimes \bPsi.
\end{equation}
Here $\bSigma_b$ can be viewed as the covariance of $\bb$ given $\bPsi$ and $\bnu$.

\subsection{Update Rules}

Minimizing (\ref{eq:BARM_cost}) is a non-convex optimization problem, and we employ standard upper bounds for this purpose leading to an EM-like algorithm.  In particular, we compute separate bounds, parameterized by auxiliary variables, for both the first and second terms of $\calL(\bPsi,\bnu)$.  While the general case can easily be handled and may be applicable for more challenging problems, here for simplicity and ease of presentation we consider minimizing $\calL(\bPsi) \triangleq \calL(\bPsi, \bnu = \mathbf{1})$, meaning all elements of $\bnu$ are fixed at one (and such is the case for all experiments reported herein, although we are currently exploring situations where this added generality could be especially helpful).

Based on \cite{wipf2011latent}, for the first term in (\ref{eq:BARM_cost}) we have
\begin{equation} \label{eq:term1_upper_bound}
\bb^{\top} \bSigma_b^{-1}  \bb \hspace*{0.2cm}  \leq \hspace*{0.2cm}  \frac{1}{\lambda}\Vert \bb - \bA \bx \Vert_2^2 + \bx^{\top} \bar{\bPsi}^{-1} \bx
\end{equation}
with equality whenever $\bx$ satisfies (\ref{eq:posterior_mean}).  For the second term we use
\begin{equation} \label{eq:term2_upper_bound}
\begin{split}
\log |\bSigma_b| \equiv  m \log|\bPsi| + \log\left| \lambda \bA^{\top} \bA  +  \bar{\bPsi}^{-1} \right| \\
\leq  m \log|\bPsi| + \mytr\left[\bPsi^{-1} \nabla_{\Psi^{-1}}\right] + C,
\end{split}
\end{equation}
where because $\log\left| \lambda \bA^{\top} \bA  +  \bar{\bPsi}^{-1} \right|$ is concave with respect to $\bPsi^{-1}$, we can upper bound it using a first-order approximation with a bias term $C$ that is independent of $\bPsi$.\footnote{If $\bPsi$ is not invertible, an effectively equivalent form of bound can nonetheless be derived.  Regardless, the final update rules do not actually depend on $\bPsi^{-1}$ anyway, and hence the algorithm can progress even as $\bPsi$ may become low rank.}  Equality is obtained when the gradient satisfies
\begin{equation} \label{eq:posterior_cov}
\nabla_{\Psi^{-1}} = \sum_{i=1}^m  \bPsi - \bPsi \bA_i^{\top} \left(  \bA \bar{\bPsi} \bA^{\top} + \lambda \bI   \right)^{-1}  \bA_i \bPsi,
\end{equation}
where $\bA_i \in \mathbb{R}^{p\times n}$ is defined such that $\bA = [\bA_1,\ldots,\bA_m]$.  Finally given the upper bounds from (\ref{eq:term1_upper_bound}) and (\ref{eq:term2_upper_bound}) with $\bX$ and $\nabla_{\Psi^{-1}}$ fixed, we can compute the optimal $\bPsi$ in closed form by optimizing the relevant $\bPsi$-dependent terms via
\begin{eqnarray} \label{eq:psi_update}
\bPsi^{opt} & = & \arg\min_{\bX} \mytr\left[  \bPsi^{-1} \left( \bX \bX^{\top} + \nabla_{\Psi^{-1}}\right) \right]  + m \log|\bPsi| \nonumber \\
 & = & \frac{1}{m} \left[\hat{\bX} \hat{\bX}^{\top} + \nabla_{\Psi^{-1}} \right].
\end{eqnarray}
By agnostically starting with $\bPsi = \bI$ and then iteratively computing (\ref{eq:posterior_mean}), (\ref{eq:posterior_cov}), and (\ref{eq:psi_update}), we can then obtain an estimate for $\bPsi$, and more importantly, a corresponding estimate for $\bX$ given by (\ref{eq:posterior_mean}) at convergence.  We refer to this basic procedure as BARM for \emph{Bayesian Affine Rank Minimization}.  The next section will detail why it is particularly well-suited for solving problems such as (\ref{eq:affine_rank_problem}).

%, as well as additional algorithmic enhancements.

% seemingly unremarkable algorithm

\section{Properties of BARM} \label{sec:analysis}
Here we first describe a close but perhaps not intuitively-obvious relationship between the BARM objective function and canonical nuclear norm minimization.  We then discuss desirable properties of global and local minima before concluding with a brief examination of convergence issues.

\subsection{Connections with Nuclear Norm Minimization} \label{sec:connection_with_NN}
On the surface, it may appear that minimizing (\ref{eq:BARM_cost}) is completely unrelated to the convex problem
\begin{equation} \label{eq:basic_nuclear norm}
\min_{\bX} \| \bX \|_* ~~\mbox{s.t.}~ \bb = \mathcal{A}(\bX)
\end{equation}
that is most commonly associated with practical rank minimization implementations.  However, a close connection can be revealed by considering the modified objective function
\begin{equation} \label{eq:dual_nuclear_norm}
\calL'(\bPsi) = \bb^{\top} \bSigma_b^{-1} \bb + \mbox{tr}[\bar{\bPsi}],
\end{equation}
which represents nothing more than (\ref{eq:BARM_cost}), with $\bnu = \mathbf{1}$ and with $\log|\bSigma_b|$ being replaced by $\mbox{tr}[\bar{\bPsi}]$.  Now suppose we minimize (\ref{eq:dual_nuclear_norm}) with respect to $\bPsi \in H^+$ obtaining some $\bPsi^*$.  We then go on to compute an estimate of $\bX$ using (\ref{eq:posterior_mean}).  Note that if we apply the bound from (\ref{eq:term1_upper_bound}) to the first term in (\ref{eq:dual_nuclear_norm}), then this estimate for $\bX$ equivalently solves
\begin{equation} \label{eq:nuclear_norm_auxiliary}
\min_{\bPsi \in H^+, \bX} \frac{1}{\lambda}\Vert \bb - \bA \bx \Vert_2^2 + \bx^{\top} \bar{\bPsi}^{-1} \bx + \mbox{tr}[\bar{\bPsi}],
\end{equation}
with $\bx = \myvec[\bX]$ as before.  If we first optimize over $\bPsi$, it is easily demonstrated that the optimal value of $\bPsi$ equals $(\bX \bX^{\top} )^{1/2}$.  Plugging this value into (\ref{eq:nuclear_norm_auxiliary}), simplifying, and then applying the definition of the nuclear norm, we arrive at
\begin{equation} \label{eq:nuclear_norm_primal}
\min_{\bX} \frac{1}{\lambda}\Vert \bb - \bA \bx \Vert_2^2 + 2 \| \bX \|_*,
\end{equation}
Furthermore, in the limit $\lambda \rightarrow 0$ (applied outside of the minimization), (\ref{eq:nuclear_norm_primal}) becomes equivalent to (\ref{eq:basic_nuclear norm}).

Consequently, we may conclude that the central distinction between the proposed BARM cost function and nuclear norm minimization is an intrinsic $\mathcal{A}$-dependent penalty function $\log|\bSigma_b|$ which is applied in covariance space.  In Section \ref{sec:minima_analysis} we will examine desirable properties of this non-convex substitution, highlighting our desire to treat the underlying BARM probabilistic model as an independent cost function that may be subject to technical analysis independent of its Bayesian origins.  For more information regarding the duality relationship between variance/covariance space and coefficient space, at least in the related context of compressive sensing models, please refer to \cite{wipf2011latent}.

\subsection{Global/Local Minima Analysis} \label{sec:minima_analysis}
As discussed in Section \ref{sec:related_work} one nice property of the $\sum_i \log \left( \sigma_i[\bX] \right)$ penalty employed (approximately) by IRLS0 \cite{mohan2012iterative} is that it can be viewed as a smooth version of the matrix rank function while still possessing the same set of minimum, both global and local, over the affine constraint set, at least if we consider the limiting situation of $\sum_i \log \left( \sigma_i[\bX]^2 + \gamma \right)$ when $\gamma$ becomes small so that we may avoid the distracting singularity of $\log 0$.  Additionally, it possesses an attractive form of scale invariance, meaning that if $\bX^*$ is an optimal feasible solution, a block-diagonal rescaling of $\bA$ nevertheless leads to an equivalent rescaling of the optimum (without the need for solving an additional optimization problem using the new $\bA$).  This is very much unlike the nuclear norm or other non-convex surrogates that penalize the singular values of $\bX$ in a scale-dependent manner.

In contrast, the proposed algorithm is based on a very different Gaussian statistical model with seemingly a more tenuous connection with rank minimization. Encouragingly however, the proposed cost function enjoys the  same global/local minima properties  as $\sum_i \log \left( \sigma_i[\bX]^2 + \gamma \right)$ with $\gamma \rightarrow 0$.  Before presenting these results, we define $\spark[\bA]$ as the smallest number of linearly dependent columns in matrix $\bA$ \cite{donoho2003optimally}.  All proofs are deferred to the Appendix.

\begin{lemma} \label{lem:global_min}
Define $r$ as the smallest rank of any feasible solution to $\bb = \bA \myvec[\bX] $, where $\bA \in \mathbb{R}^{p\times nm}$ satisfies $\spark[\bA] = p+1$.  Then if  $r < p/m$, any global minimizer $\{\bPsi^*,\bnu^*\}$ of (\ref{eq:BARM_cost}) in the limit $\lambda \rightarrow 0$ is such that $\bx^* = \bar{\bPsi}^* \bA^{\top} \left( \bA \bar{\bPsi}^* \bA^{\top}\right)^{\dag} \bb$ is feasible and $\rank[\bX^*] = r$ with $\myvec[\bX^*] = \bx^*$.
\end{lemma}
\begin{lemma} \label{lem:scale_invariance}
Additionally, let $\tilde{\bA} = \bA D$, where $\bD = \mydiag[\alpha_1\bGamma, \ldots,\alpha_m\bGamma]$ is a block-diagonal matrix with invertible blocks $\bGamma \in \mathbb{R}^{n \times n}$ of unit norm scaled with coefficients $\alpha_i > 0$.  Then iff $\{\bPsi^*,\bnu^*\}$ is a minimizer (global or local) to (\ref{eq:BARM_cost}) in the limit $\lambda \rightarrow 0$, then $\{\bGamma^{-1}\bPsi^*, \mydiag[\balpha]^{-1} \bnu^*\}$ is a minimizer when $\tilde{\bA}$ replaces $\bA$.  The corresponding estimates of $\bX$ are likewise in one-to-one correspondence.
\end{lemma}

\textbf{\emph{Remarks:}}   The assumption $r = \rank[\bX^*] < p/m$ in Lemma \ref{lem:global_min} is completely unrestrictive, especially given that a unique, minimal-rank solution is only theoretically possible by \emph{any} algorithm if $p \geq (n+m)r-r^2$, which is much more restrictive than $p>r m$. Hence the bound we require is well above that required for uniqueness anyway.  Likewise the spark assumption will be satisfied for any $\bA$ with even an infinitesimal (continuous) random component.  Consequently, we are essentially always guaranteed that BARM possesses the same global optimum as the rank function.  Regarding Lemma \ref{lem:scale_invariance},  no surrogate rank penalty of the form $\sum_{i} f(\sigma_i[\bX])$ can achieve this result except for $f(z) = \log z$, or inconsequential limiting translations and rescalings of the $\log$ such as the indicator function $I[z \neq 0]$ (which is related to the log via arguments in Section \ref{sec:related_work}).

While these results are certainly a useful starting point, the real advantage of adopting the BARM cost function is that locally minimizing solutions are exceedingly rare, largely as a consequence of the marginalization process in (\ref{eq:marginal_like}), and in some cases provably so.  A specialized example of this smoothing can be quantified in the following scenario.

Suppose $\bA$ is now block diagonal, with diagonal blocks $\bA_i$ such that $\bb_i = \bA_i \bx_{:i}$ producing the aggregate observation vector $\bb = [\bb_1^{\top},\ldots,\bb_m^{\top}]^{\top}$.  While somewhat restricted, this situation nonetheless includes many important special cases, including canonical matrix completion and generalized matrix completion where elements of $\bZ_0 \triangleq \bW \bX_0$ are observed after some transformation $\bW$, instead of $\bX_0$ directly.

\begin{theorem} \label{thm:local_min_theorem}
Let $\bb = \bA \myvec[\bX] $, where $\bA$ is block diagonal, with blocks $\bA_i \in \mathbb{R}^{p_i \times n}$.  Moreover, assume $p_i > 1$ for all $i$ and that $\cap_{i} \mbox{null}[\bA_i] = \emptyset$.  Then if $\min_{\bX} \rank[\bX] = 1$ in the feasible region, any minimizer $\{\bPsi^*,\bnu^*\}$ of (\ref{eq:BARM_cost}) (global or local) in the limit $\lambda \rightarrow 0$ is such that $\bx^* = \bar{\bPsi}^* \bA^{\top} \left( \bA \bar{\bPsi}^* \bA^{\top}\right)^{\dag} \bb$ is feasible and $\rank[\bX^*] = 1$ with $\myvec[\bX^*] = \bx^*$.  Furthermore, no cost function in the form of (\ref{eq:general_low_rank})
can satisfy the same result.  In particular, there can always exist local and/or global minima with rank greater than one.
\end{theorem}

\textbf{\emph{Remarks:}} This result implies that, under extremely mild conditions, which do not even depend on the concentration properties of $\bA$, the proposed cost function has no minima that are not global minima.  (The minor technical condition regarding nullspace intersections merely ensures that high-rank components cannot simultaneously ``hide" in the nullspace of every measurement matrix $\bA_i$; the actual $\bA$ operator may still be highly ill-conditioned.)  Thus any algorithm with provable convergence to some local minimizer is guaranteed to obtain a globally optimal solution.\footnote{Note also that with minimal additional effort, it can be shown that no suboptimal stationary points of any kind, including saddle points, are possible.}

Interestingly, such a guarantee is not possible with any other penalty function of the standard form $\sum_{i} f(\sigma_i[\bX])$, which is the typical recipe for rank minimization algorithms, convex or not.  Additionally, if a unique rank-one solution exists to (\ref{eq:affine_rank_problem}), then the unique minimizing solution to (\ref{eq:BARM_cost}) will produce this $\bX$ via (\ref{eq:posterior_mean}).  Crucially, this will occur even when the minimal number of measurements $p = n+m-1$ are available, unlike any other algorithm we are aware of that is blind to the true underlying rank.\footnote{It is important to emphasize that the difficulty of estimating the optimal low-rank solution is based on the ratio of the d.o.f. in $\bX$ to the number of observations $p$.  Consequently, estimating $\bX$ even with $r$ small can be challenging when $p$ is also small, meaning $\bA$ is highly overcomplete.}  And importantly, the underlying intuition that local minima are smoothed away nonetheless carries over to situations where the rank is greater than one. %The supplementary file contains an enlightening visualization of this smoothing effect.

\subsection{Visualization of BARM Local Minima Smoothing}
\label{ssec:visual}

%\textbf{\emph{Visualization:}}
To further explore the smoothing effect and complement Theorem 1,  it helps to visualize rank penalty functions restricted to the feasible region.  While the BARM algorithm involves minimizing (\ref{eq:BARM_cost}), its implicit penalty function on $\bX$ can nonetheless be numerically obtained across the feasible region in a given subspace of interest; for other penalties such as the nuclear norm this is of course trivial.  Practically it is convenient to explore a 1D feasible subspace generated by $\bX^* + \eta \bV$, where $\bX^*$ is the true minimum rank solution, $\bV \in \mbox{null}[\bA]$, and $\eta$ is a scalar.  We may then plot various penalty function values as $\eta$ is varied, tracing the corresponding 1D feasible subspace.  We choose $\bV = \bX^1 - \bX^*$, where $\bX^1$ is a feasible solution with minimum nuclear norm; however, random selections from $\mbox{null}[\bA]$ also show similar characteristics.

\begin{figure*}
    \centering
    \includegraphics[width = 1.9\columnwidth]{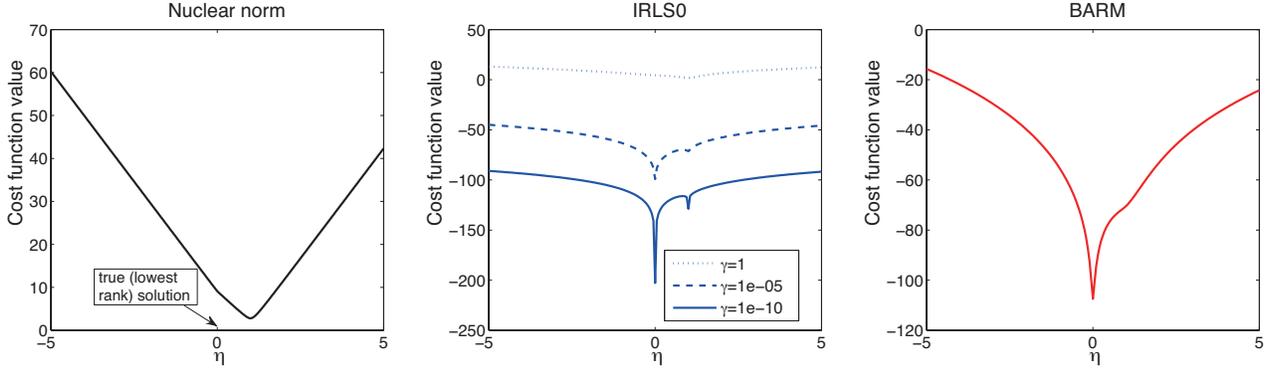}
    \caption{\small{Plots of different surrogates for matrix rank in a 1D feasible subspace. Here the convex nuclear norm does not retain the correct global minimum.  In contrast, although the non-convex $\sum_i \log \left(\sigma_i[\bX]^2 + \gamma \right)$ penalty exhibits the correct minimum when $\gamma$ is sufficiently small, it also contains spurious minima.  Only BARM smoothes away local minimum while simultaneously retaining the correct global optima.}}
    \label{fig:localsmooth}
    %\vspace{-3mm}
\end{figure*}

Figure \ref{fig:localsmooth} provides a simple example of this process.  $\bA$ is generated with all zeros and a single randomly placed '1' in each row leading to a canonical matrix completion problem.  $\bX^* \in \mathbb{R}^{5\times5}$ is randomly generated as $\bX^* = \bu \bv^{\top}$, where $\bu$ and $\bv$ are iid $\mathcal{N}(0,1)$ vectors, and so $\bX^*$ is rank one.  Finally, $p = 10$ elements are observed, and therefore $\bA$ has $10$ rows and $5 \times 5 = 25$ columns.  $\eta$ is varied from $-5$ to $5$ and the values of the nuclear norm, $\sum_i \log \left( \sigma_i[\bX]^2 + \gamma \right)$, and the implicit BARM cost function are displayed.

From the figure we observe that the minimum of the nuclear norm is not produced when the rank is smallest, which occurs when $\eta = 0$; hence the convex cost function fails for this problem.  Likewise, the $\sum_i \log \left( \sigma_i[\bX]^2 + \gamma \right)$ penalty used by IRLS0 displays an incorrect global minimum when the tuning parameter $\gamma$ is large.  In contrast, when $\gamma$ is small, while the global minimum may now be correct, spurious local ditches have appeared in the cost function.\footnote{Technically speaking, these are not provably local minima since we are only considering a 1D subspace of the feasible region.  However, it nonetheless illustrates the strong potential for troublesome local minima, especially in high dimensional practical problems.}  Therefore, any success of the IRLS0 algorithm depends heavily on a carefully balanced decaying sequence of $\gamma$ values, with the hope that initial iterations can steer the trajectory towards a desirable basin of attraction where local minima are less problematic.  One advantage of BARM then is that it is parameter free in this respect and yet still retains the correct global minimum, often without additional spurious local minima.

\subsection{Convergence}
\label{ssec:converge}
% \textbf{\emph{Visualization:}}

%\textbf{\emph{Convergence:}}
Previous results of Section \ref{sec:analysis} are limited to exploring aspects of the underlying BARM cost function.  Regarding the BARM algorithm itself, by construction the updates generated by (\ref{eq:posterior_mean}), (\ref{eq:posterior_cov}), and (\ref{eq:psi_update}) are guaranteed to reduce or leave unchanged $\calL(\bPsi)$ at each iteration.  However, this is not technically sufficient to guarantee convergence to a stationary point of the cost function unless, for example, the additional conditions of Zangwill's Global Convergence Theorem are satisfied \cite{zangwill1969nonlinear}.  However, provided we add a small regularization factor $\gamma \mytr[\bPsi^{-1}]$, with $\gamma > 0$ to the BARM objective, then it can be shown that any cluster point of the resulting sequence of iterations $\{\bPsi^k\}$ must be a stationary point.  Moreover, because the sequence is bounded, there will always exist at least one cluster point, and therefore the algorithm is guaranteed to at least converge to a set of parameters values $\calS$ such that for any $\bPsi^* \in \calS$, $\calL(\bPsi^*) + \gamma \mytr[(\bPsi^*)^{-1}]$  is a stationary point.

Finally, we should mention that this extra $\gamma$ factor is akin to the homotopy continuation regularizer used by the IRLS0 algorithm \cite{mohan2012iterative} as discussed in Section \ref{sec:related_work}.  However, whereas IRLS0 requires a carefully-chosen, decreasing sequence $\{\gamma^k\}$ with $\gamma^k>0$ \emph{both} to prove convergence  \emph{and} to avoid local minimum (and without this factor the algorithm performs very poorly in practice), for BARM a small, fixed factor only need be included as a technical necessity for proving formal convergence; in practice (and in our experiments) it can be fixed to zero.

\section{Symmetrization Improvements}
\label{sec:symm}

%\textbf{\emph{Symmetrization Improvements:}}
Despite the promising theoretical attributes of BARM from the previous section, there remains one important artifact of its probabilistic origins not found in more conventional existing rank minimization algorithms.  In particular, other algorithms rely upon a symmetric penalty function that is independent of whether we are working with $\bX$ or $\bX^{\top}$.  All methods that reduce  to (\ref{eq:general_low_rank}) fall into this category, e.g., nuclear norm minimization, IRNN, or IRLS0.  In contrast, our method relies on defining a distribution with respect to the columns of $\bX$.  Consequently the underlying cost function is not identical when derived with respect to $\bX$ or $\bX^{\top}$, a difference which will depend  on $\bA$.  While globally optimal solutions should nonetheless be the same, the convergence trajectory could depend on this distinction leading to different local minima in certain circumstances.  Although either construction leads to low-rank solutions, we may nonetheless expect improvement if we can somehow symmetrize the algorithm formulation.

To accomplish this, we consider a Gaussian prior on $\bx = \myvec[\bX]$ with a covariance formed using a block-wise averaging of covariances defined over rows and columns, denoted $\bPsi_r$ and $\bPsi_c$ respectively.  The overall covariance is then given by the Kronecker sum
\begin{equation}
\bar{\bPsi} = 1/2\left( \bPsi_r \otimes \bI +  \bI \otimes \bPsi_c \right).
\end{equation}
The estimation process proceeds in a similar fashion as before but with modifications and alternate upper-bounds that accommodate for this merger.  For reported experimental results this symmetric version of BARM is used, with complete update rules listed in the Appendix and computational complexity evaluated in Section \ref{sec:complexity}.

% a discussion of computational complexity in

%Add modified update rules ...
%This then leads to the following augmented udpate rules for BARM:

\section{Experimental Validation}
\label{sec:expr}

This section compares BARM with existing state-of-the-art affine rank minimization algorithms.  For BARM, in all noiseless cases we simply used $\lambda = 10^{-10}$ (effectively zero, the exact value is not important), and hence no tuning parameters are required.  Likewise, nuclear norm minimization \cite{candes2009exact,zhang2012tilt} requires no tuning parameters  beyond implementation-dependent control parameters frequently used to enhance convergence speed (however the global minimum is unaltered given that the problem is convex).  For the IRLS0 algorithm, we used our own implementation as the algorithm is straightforward and no code was available for the case of general $\mathcal{A}$; we based the required decreasing $\gamma_k$ sequence on suggestions from \cite{mohan2012iterative}.  IRLS0 code is available from the original authors for matrix completion; however, the results obtained with this code are not better than those obtained with our version.

For the IRNN algorithm of \cite{lu2014generalized}, we did not have access to code for general $\mathcal{A}$, nor specific details of how various parameters should be set in the general case.  Note also that IRNN has multiple parameters to tune even in noiseless problems unlike BARM.  Therefore we report results directly from \cite{lu2014generalized} where available.  Additionally, we emphasize that both \cite{lu2014generalized} and \cite{mohan2012iterative} show superior results to a number of other algorithms; we do not generally compare with these others given that they are likely no longer state-of-the-art and may clutter the presentation.

Moreover, we show limited empirical results with the variational sparse Bayesian algorithm (VSBL) from \cite{babacan2012sparse} because of its Bayesian origins, although the underlying parameterization is decidedly different from BARM.  But these results are limited to matrix completion as VSBL does not presently handle general affine constraints. Results from VSBL were obtained using publicly available code from the authors.

Although our focus here is on algorithms that do not require knowledge of the true rank of the optimal solution,  we have nonetheless conducted numerous experiments with \cite{jain2013low} or the normalized hard thresholding algorithm from \cite{tanner2013normalized}. Even when the correct rank is provided, results are far inferior to BARM. To avoid clutter the presentation, we address the comparison of BARM to these algorithms separately in Section \ref{ssec:com2alter}.

\begin{figure}
    \centering
    \includegraphics[width = 0.95\columnwidth]{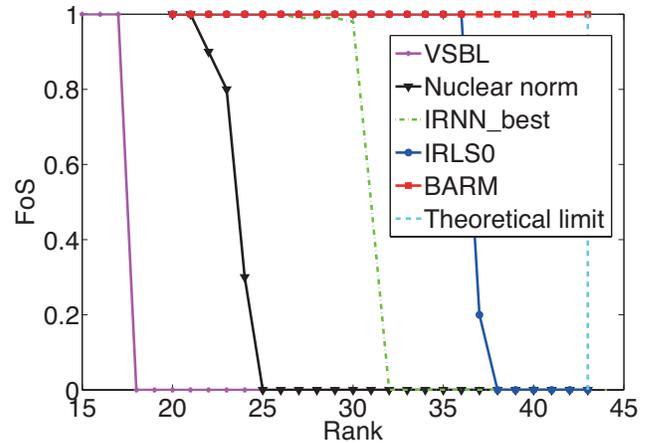}
    \caption{\small{Matrix completion comparisons (avg of 10 trials)}}
    \label{fig:mc}
\end{figure}

\subsection{Matrix Completion}
\label{ssec:mc}

%, in part because this allows us to compare our results with the latest algorithms even when code is not available

\begin{table}[t]
\caption{Matrix completion results of BARM with IRLS0 on the three hardest problems from \cite{mohan2012iterative}. Published results in \cite{mohan2012iterative} included for comparison.}
\label{tab:jmlr}
\begin{center}
\begin{tabular}{p{13pt}p{21pt}p{12pt}|p{20pt}p{16pt}p{20pt}p{20pt}|p{25pt}}
\hline
 \multicolumn{3}{p{2pt}}{Problem} & IRLS0 & IHT & FPCA& Opts & BARM \\
\hline
 FR & n(=m)  & r & FoS & FoS  & FoS  & FoS & FoS  \\
\hline
 0.78  & 500 & 20 & 0.9 & 0 & 0 & 0 & 1  \\
 0.8  & 40 & 9 & 1 & 0 & 0.5 & 0 & 1  \\
 0.87  & 100 & 14 & 0.5 & 0 & 0 & 0 & 1  \\
\hline
\end{tabular}
\end{center}
\end{table}

%\textbf{\emph{Matrix Completion:}}
We begin with the matrix completion problem from (\ref{eq:mc}).  For this purpose we reproduce the exact same experiment from \cite{lu2014generalized}, where a rank $r$ matrix is generated as $\bX_0 = \mathbf{M}_L \mathbf{M}_R$, with $\mathbf{M}_L \in \mathbb{R}^{n \times r}$  and $\mathbf{M}_R \in \mathbb{R}^{r \times m}$ ($n=m=150$) as iid $\calN(0,1)$ random matrices.  50\% of all entries are then hidden uniformly at random.  The \emph{relative error} (REL) is defined by $\| \bX_0 - \hat{\bX} \|_{\calF}/\| \bX_0 \|_{\calF}$ for each trial.  The \emph{frequency of success} (FoS) score,  which measures the percentage of trials where the REL is below $10^{-3}$, is then computed and averaged across trials as $r$ is varied.  Results are shown in Figure \ref{fig:mc} where BARM is the only algorithm capable of reaching the theoretical recovery limit, beyond which $p = 0.5 \times 150^2 = 11250$ is surpassed by the number of degrees of freedom in $\bX_0$, in this case $2\times150\times44 - 44^2 = 11264$.  Note that FoS values were reported in \cite{lu2014generalized} over a wide range of non-convex IRNN algorithms.  The green curve represents the best performing candidate from this pool as tuned by the original authors.  Interestingly, although VSBL is based on a probabilistic model as is BARM, the underlying parameterization, cost function, and update rules are entirely different and do not benefit from any strong theoretical underpinnings.  Hence performance does not always match recent state-of-the-art algorithms.

\begin{figure*}
	\centering
    \subfigure[$50 \times 50$, $\bA$ uncorrelated]
    {
        \label{fig:amrm50}
        \includegraphics[width = 0.96\columnwidth]{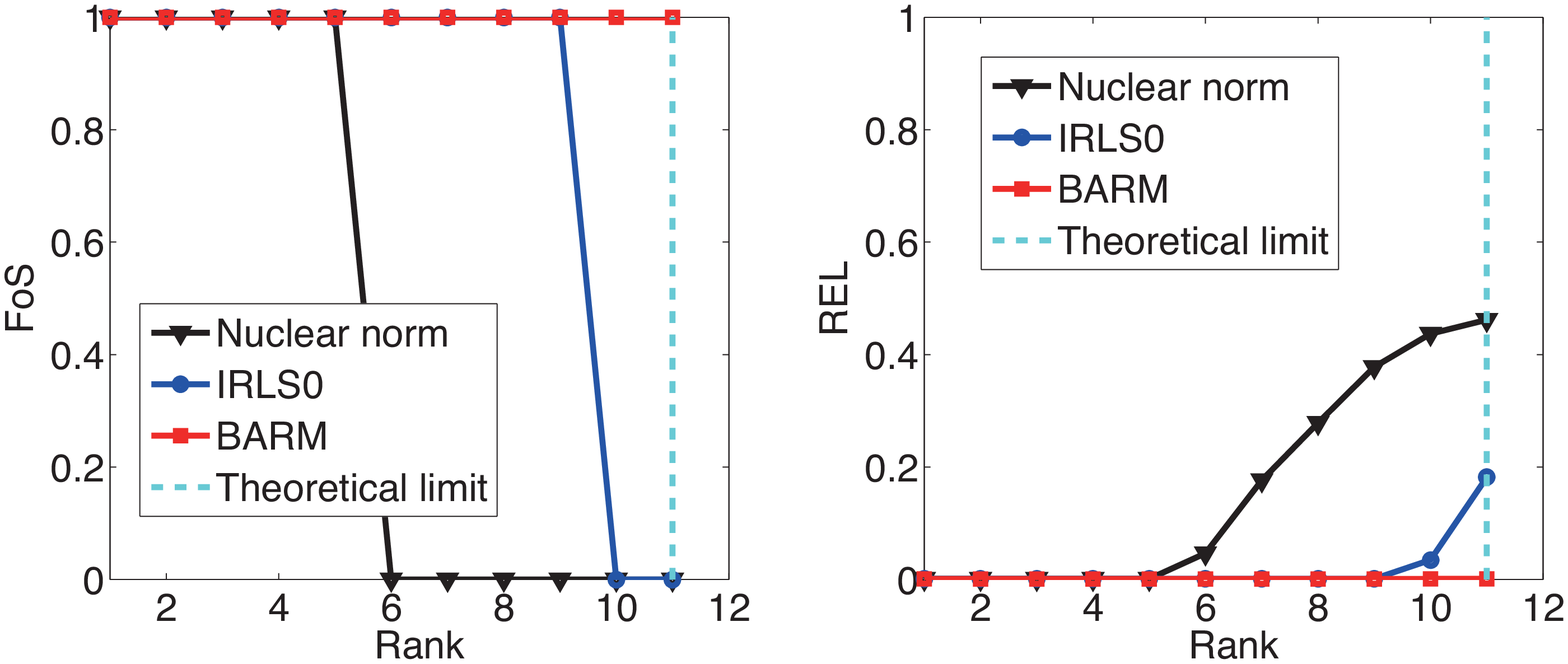}
    }
    \subfigure[$50 \times 50$, $\bA$ correlated]
    {
        \label{fig:amrm100}
        \includegraphics[width = 0.96\columnwidth]{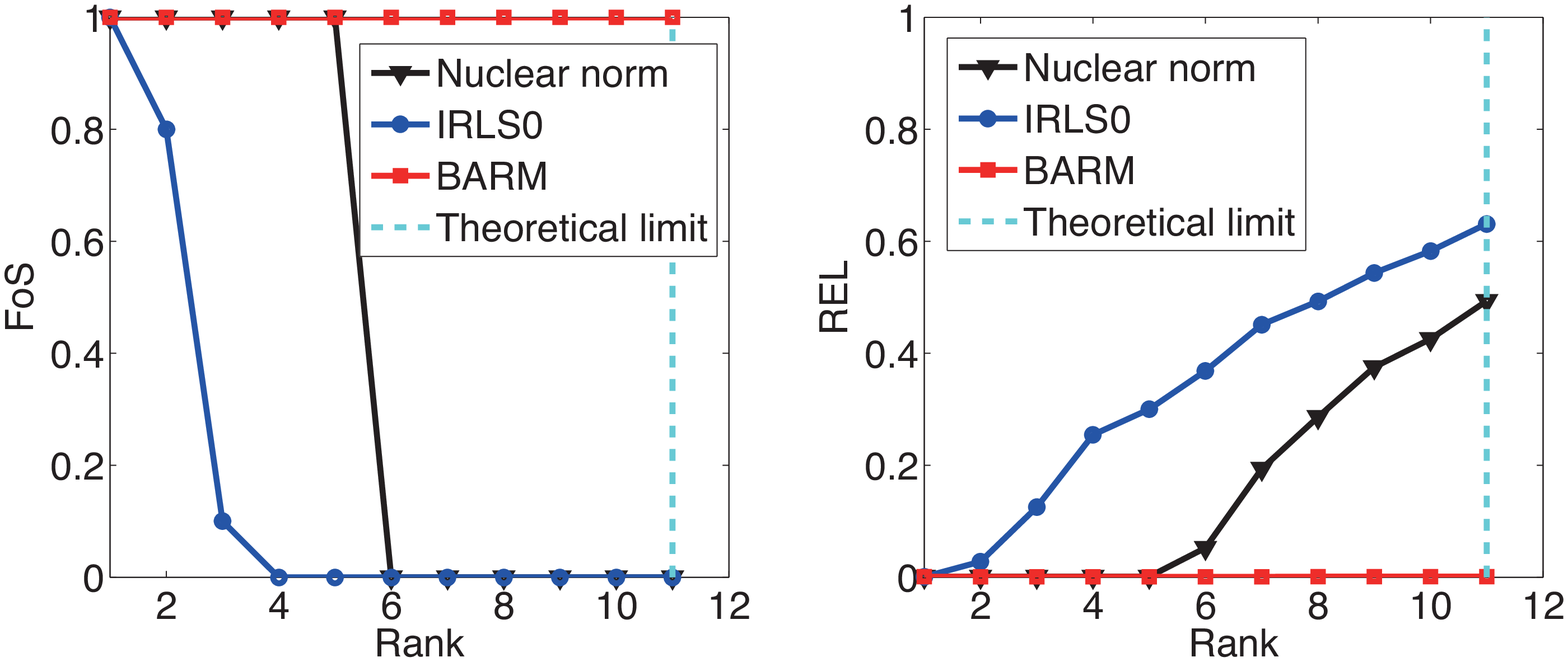}
    }
    \vspace{-2mm}
    \subfigure[$100 \times 100$, $\bA$ uncorrelated]
    {
        \label{fig:amrm50}
        \includegraphics[width = 0.96\columnwidth]{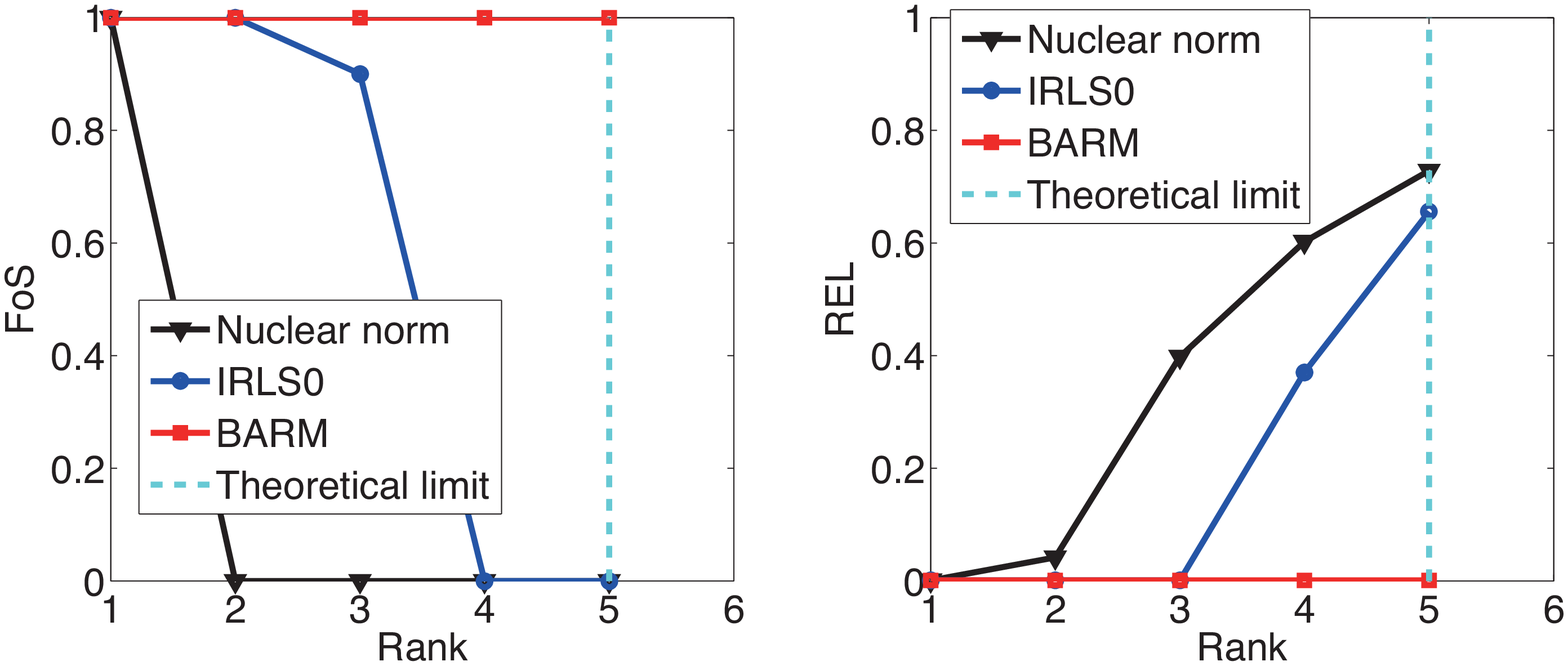}
    }
    \subfigure[$100 \times 100$, $\bA$ correlated]
    {
        \label{fig:amrm100}
        \includegraphics[width = 0.96\columnwidth]{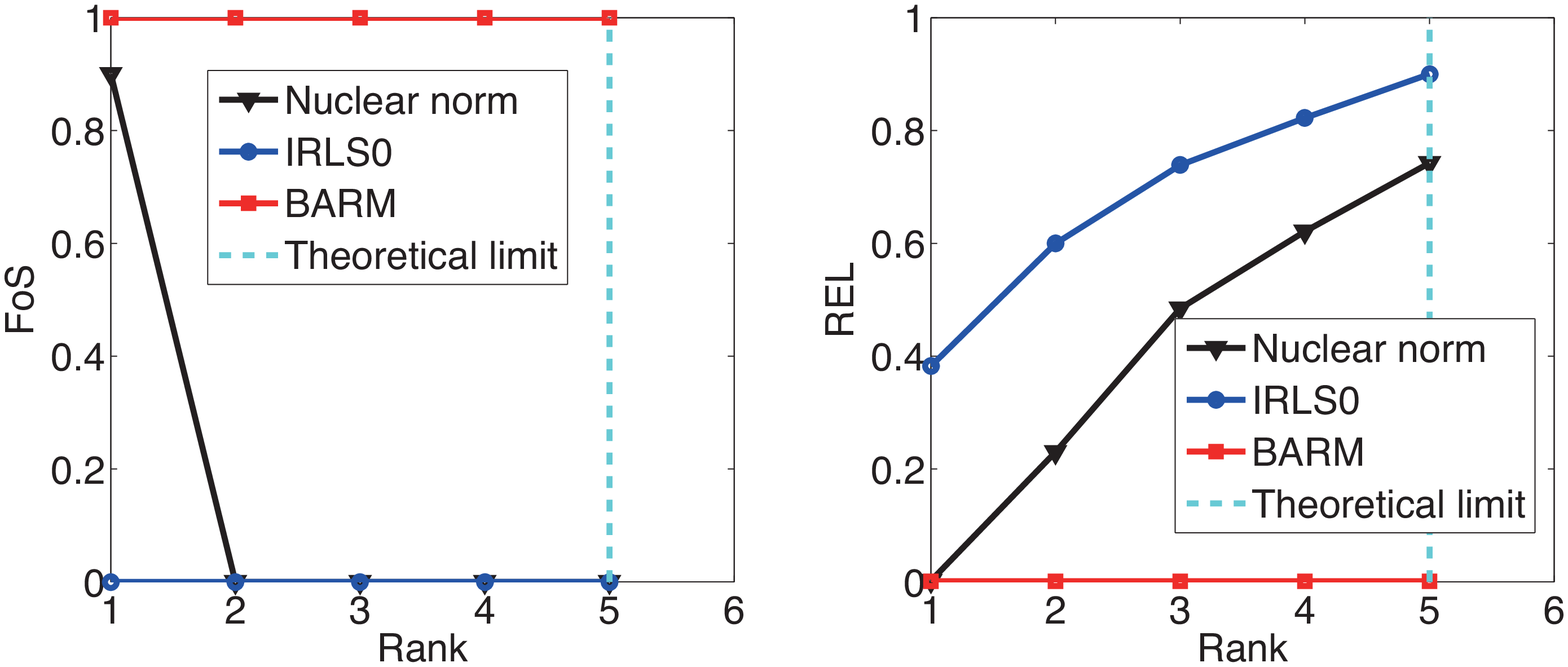}
    }
	\caption{\small{Comparisons with general affine constraints (avg of 10 trials)}}
\label{fig:amrm}
\end{figure*}
% so as to provide direct head-to-head comparisons with the authors' original implementation and settings.

Besides BARM, the IRLS0 algorithm also displayed better performance than the other methods.  This motivated us to reproduce some of the matrix completion experiments from \cite{mohan2012iterative} which presumably were designed to showcase difficult regimes where IRLS0 is superior.  For this purpose, $\bX_0$ is conveniently generated in the same way as above; however, values of $n$, $m$, $r$, and the percentage of missing entries are varied while evaluating reconstructions using FoS.  While \cite{mohan2012iterative} tests a variety of combinations of these values to explore varying degrees of problem difficulty, here we only reproduce the most challenging cases to see if BARM is still able to produce superior reconstruction accuracy.  In this respect problem difficulty is measured by the \emph{degrees of freedom ratio} (FR) given by FR$=r(n+m-r)/p$ as defined in \cite{mohan2012iterative}. We also only include experiments where algorithms are blind to the true rank of $\bX_0$.\footnote{Note that IRLS0 can be modified to account for the true rank if such knowledge were available.}  Results are shown in Table \ref{tab:jmlr}, where we have also displayed the published results both IRLS0 and three additional algorithms that were previously evaluated in \cite{mohan2012iterative}, namely, IHT \cite{jain2010guaranteed}, FPCA \cite{goldfarb2011convergence} and Optspace \cite{keshavan2009gradient}.  From the table we observe that, in the most difficult problem considered in \cite{mohan2012iterative}, IRLS0 achieved only a 0.5 FoS score (meaning failure 50\% of the time) while BARM still achieves a perfect 1.0.

\subsection{General $\bA$}
\label{ssec:gen}

%\textbf{\emph{General $\bA$:}}
Next we consider the more challenging problem involving arbitrary affine constraints using the implementations we had available for nuclear norm minimization, IRLS0, and BARM.  The desired low-rank $\mathbf{X}_0$ is generated in the same way as above.  We then consider two types of linear mappings where $\bA$ is generated as: (i) an iid $\mathcal{N}(0,1)$, $p \times n^2$ matrix, and (ii) $\sum_{i=1}^{p} i^{-1/2} \bu_i \bv_i^{\top}$, where $\bu_i \in \mathbb{R}^{p}$ and $\bv_i \in \mathbb{R}^{n^2}$ are iid $\mathcal{N}(0,1)$ vectors.  The latter is meant to explore less-than-ideal conditions where the linear operator displays correlations and may be somewhat ill-conditioned.  Figure \ref{fig:amrm} displays aggregate results when $\bX_0$ is $50\times 50$ and $100\times 100$, including the underlying REL scores for additional comparison.  In both cases $p=1000$ observations are used, and therefore the corresponding measurement matrices $\bA$ are $1000 \times 2500$ and $1000 \times 10000$ respectively.  We then vary $r$ from 1 up to the theoretical limit corresponding to problem size.  Again we observe that BARM is consistently able to work up to the limit, even when the $\bA$ operator is no longer an ideal Gaussian.  In general, we have explored a wide range of empirical conditions too lengthly to report here, and it is only very rarely, and always near the theoretical boundary, where BARM occasionally may not succeed.  We explore such failure cases in the next section.

\subsection{Failure Case Analysis}
\label{ssec:fail}

%\textbf{\emph{Failure Case Analysis:}}
Thus far we have not shown any cases where BARM actually fails. Of course solving (\ref{eq:affine_rank_problem}) for general $\bA$ is NP-hard so recovery failures certainly must exist in some circumstances when using a polynomial-time algorithm such as BARM.  Although we certainly cannot explore every possible scenario, it behooves us to probe more carefully for conditions under which such errors may occur.  One way to accomplish this is to push the problem difficulty even further towards the theoretical limit by reducing the number of measurements $p$ as follows.

With the number of observations fixed at $p=1000$ and a general measurement matrix $\bA$, the previous section examined the recovery of $50\times50$ and $100\times100$ matrices as the rank was varied from 1 to the recovery limit ($r = 11$ for the $50\times50$ case; $r=5$ for the $100\times 100$ case).  However, it is still possible to make the problem even more challenging by fixing $r$ at the limit and then reducing $p$ until it exactly equals the degrees of freedom $2n^2-r^2$.  With $\{n=50, r=11\}$ this occurs at $p=979$, for $\{n=100,r=5\}$ this occurs at $p=975$.

\begin{figure}
	\centering
    \subfigure[$50 \times 50$]
    {
        \label{fig:amrm50}
        \includegraphics[width = 0.45\columnwidth]{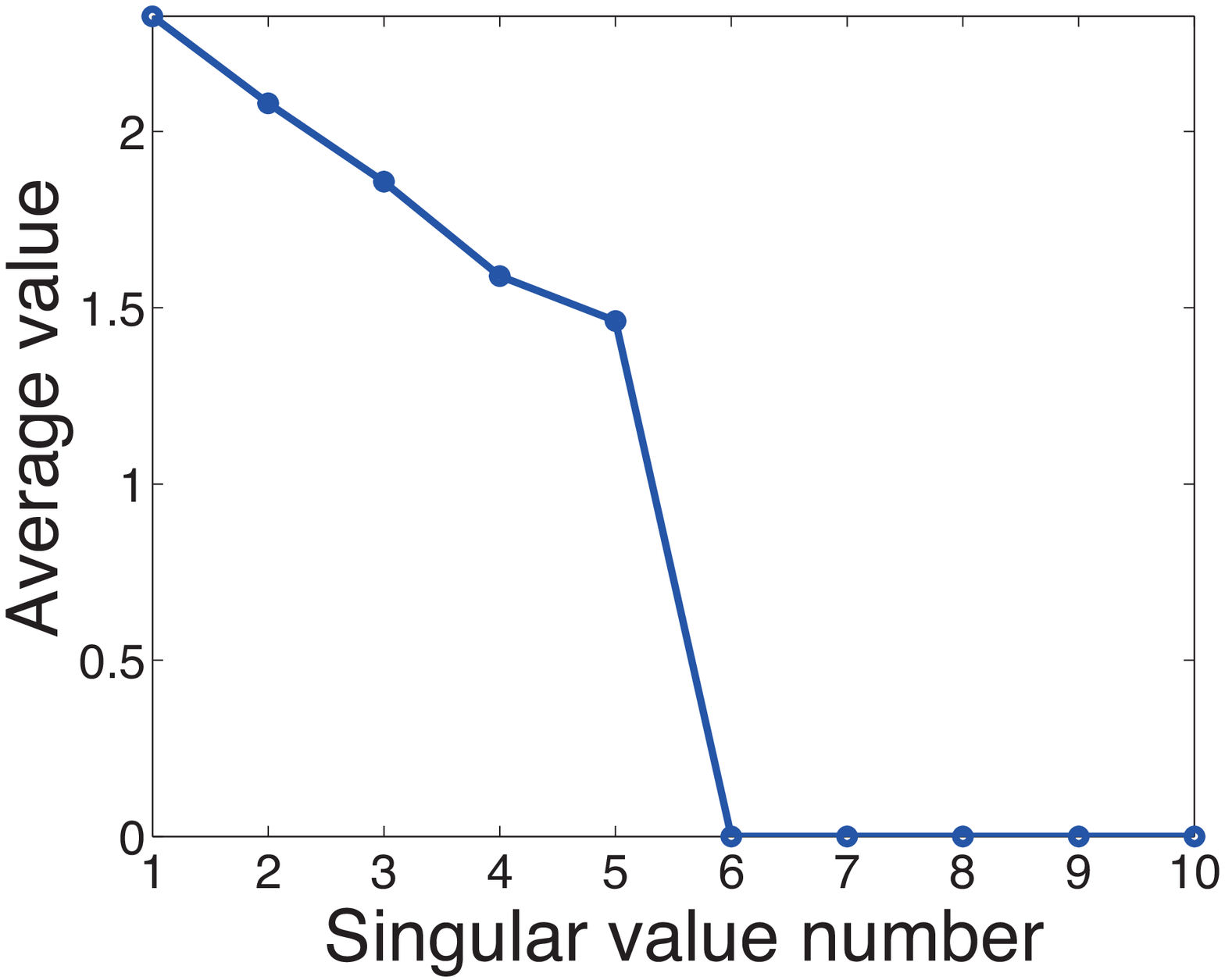}
    }
    \subfigure[$100 \times 100$]
    {
        \label{fig:amrm100}
        \includegraphics[width = 0.45\columnwidth]{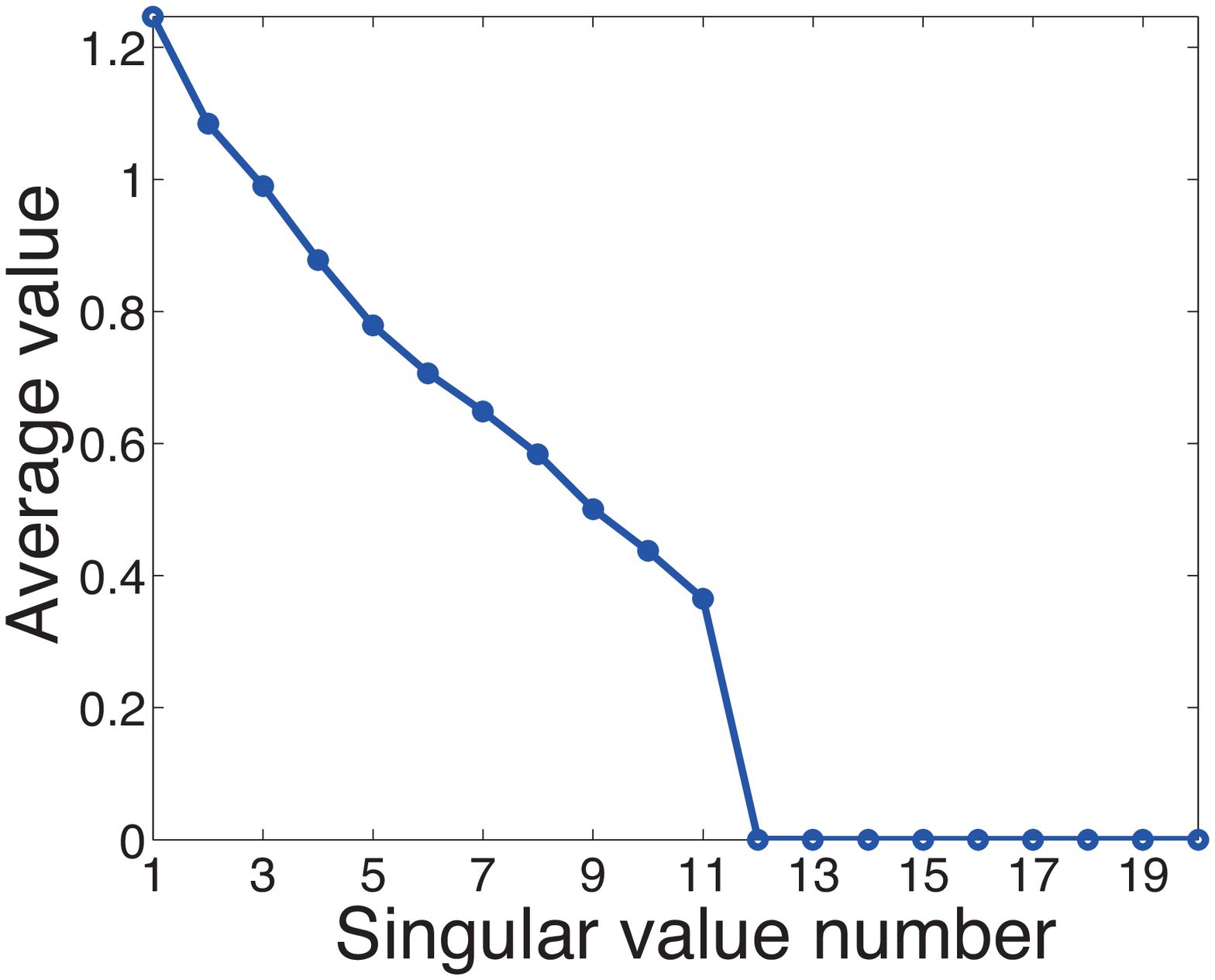}
    }
    \caption{\small{Singular value averages of failure cases.  In both cases solutions of minimal rank are obtained even though $\hat{\bX} \neq \bX_0$.}}
\label{fig:fail}
%\vspace{-10pt}
\end{figure}

We examined the BARM algorithm under these conditions with 10 additional trials using the uncorrelated $\bA$ for each problem size.  Encouragingly, BARM was still 30\% successful with $\{n=50, r=11\}$,  and 40\% successful with $\{n=100, r=5\}$.  However, it is interesting to further examine the nature of these failure cases.  In Figure \ref{fig:fail} we have averaged the singular values of $\hat{\bX}$ in all the failure cases.  Here we notice that, although the recovery was technically classified as a failure since the relative error (REL) was above the stated threshold, the estimated matrices are of almost exactly the correct minimal rank.  Hence BARM has essentially uncovered an alternative solution with minimal rank that is nonetheless feasible by construction.  We therefore speculate that right at the theoretical limit, when $\bA$ is maximally overcomplete ($p \times n^2 = 979 \times 2500$ or $975 \times 10000$ for the two problem sizes), there exists multiple feasible matrices with singular value spectral cut-off points indistinguishable from the optimal solution.  Importantly, when the other algorithms we tested failed, the failure is much more dramatic and a clear spectral cut-off at the correct rank is not apparent.

This motivates a looser success criteria than FoS to account for the possibility of multiple (nearly) optimal solutions that may not necessarily be close with respect to relative error.  For this purpose we define the \emph{frequency of rank success} (FoRS) as the percentage of trials whereby a feasible solution $\hat{\bX}$ is found such that $\sigma_r[\hat \bX]/\sigma_{r+1}[\hat \bX]>10^3$, where $\sigma_i[\cdot]$ denotes the $i$-th singular value of a matrix and $r$ is the rank of the true low-rank $\bX_0$.  In words, FoRS measures the percentage of trials such that roughly a rank $r$ solution is recovered, regardless of proximity to $\bX_0$.

Under this new criteria, all of the failure cases with respect to FoS described above, for both problem sizes, become successes; however, none of the other algorithms show improvement under this criteria, indicating that their original failures involved actual sub-optimal rank solutions.  Something similar happens when we revisit the matrix completion experiments.  For example,  based on Table \ref{tab:jmlr} the most difficult case involves FR$=0.87$; however, by further reducing $p$, we can push FR towards $1.0$ to further investigate the break-down point of BARM.  Results are shown in Table \ref{tab:jmlr2}.  While IRLS0 (which is the top performing algorithm in \cite{mohan2012iterative} and in our experiments besides BARM) fails 100\% of the time via both metrics, BARM can achieve an FoS of $0.7$ even when FR$ = 0.99$ and an FoRS of $1.0$ in all cases.

\begin{figure*}
	\centering
    \subfigure[$50 \times 50$, $\bA$ sub-sampled DCT]
    {
        \label{fig:amrm50}
        \includegraphics[width = 0.96\columnwidth]{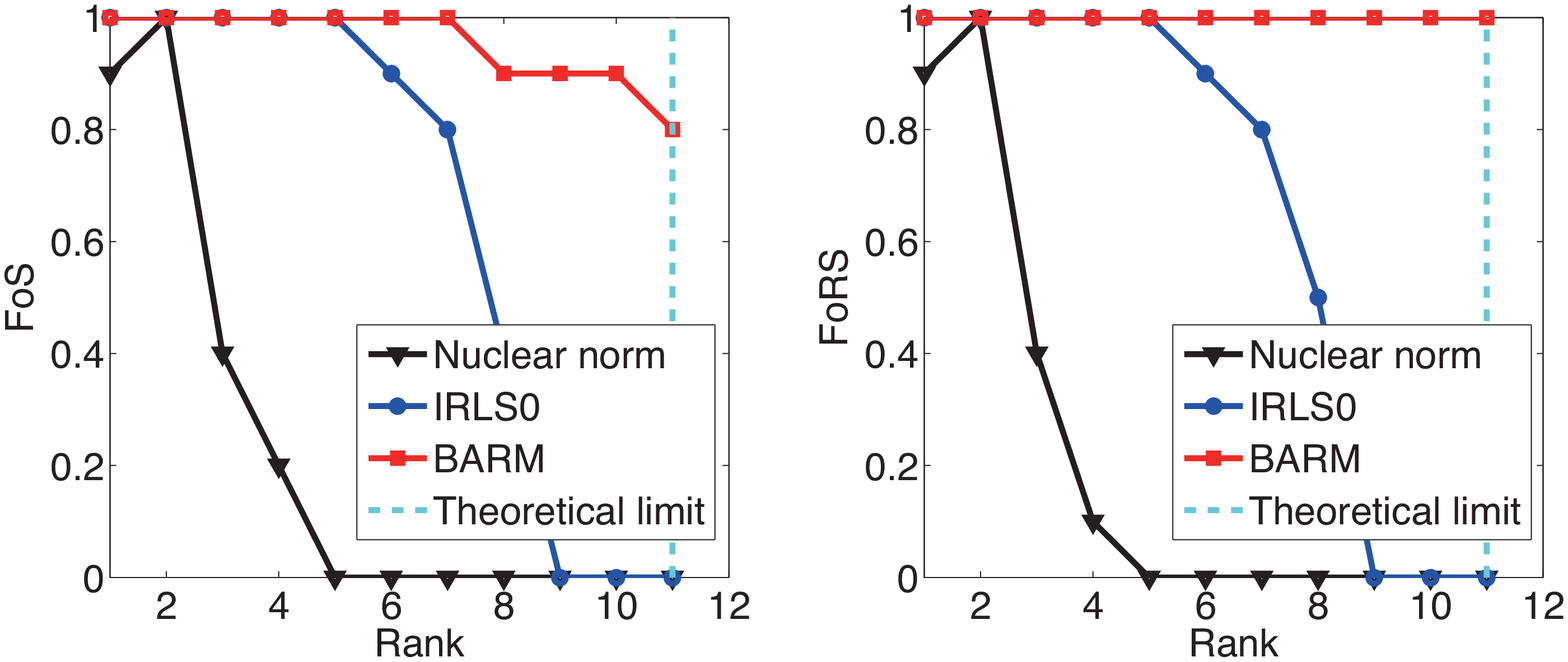}
    }
    \subfigure[$100 \times 100$, $\bA$ sub-sampled DCT]
    {
        \label{fig:amrm100}
        \includegraphics[width = 0.96\columnwidth]{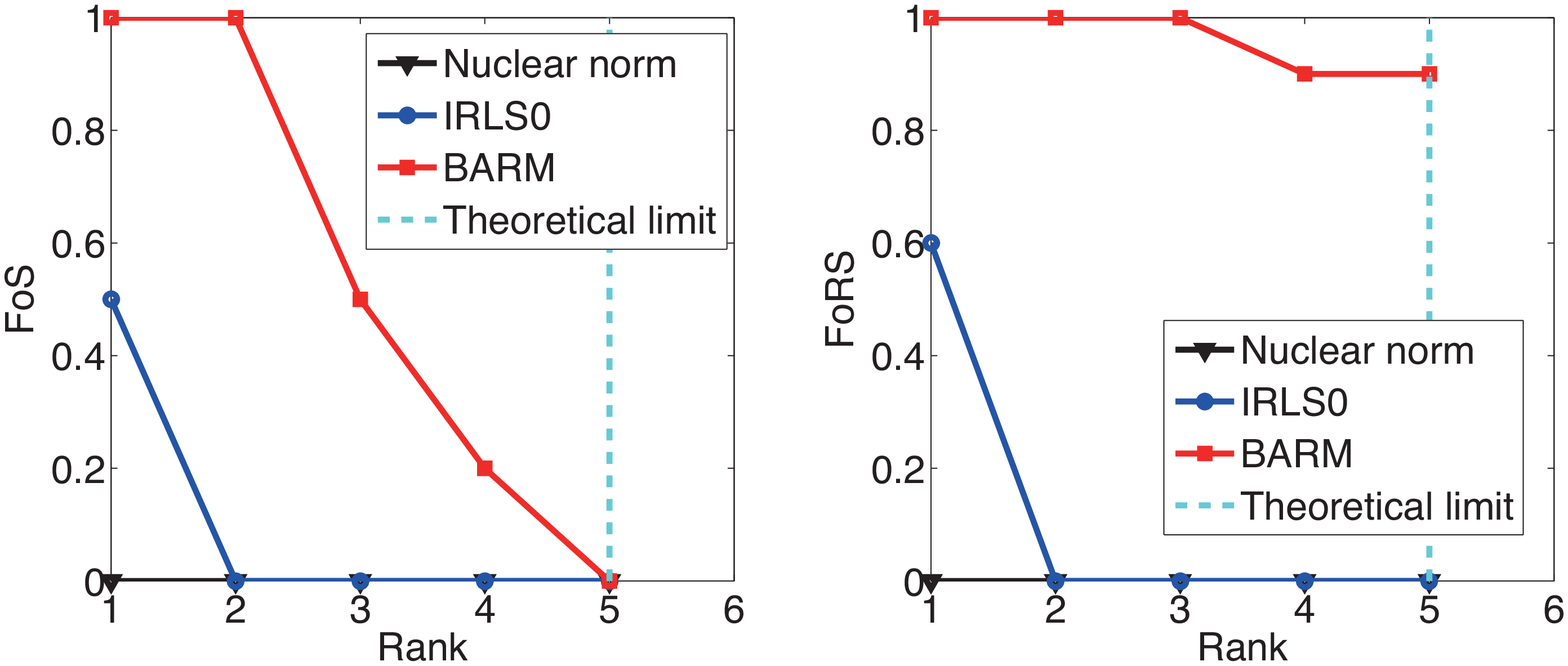}
    }
	\caption{\small{Comparisons with structured affine constraints using both FoS and FoFS evaluation metrics (avg of 10 trials).}}
\label{fig:dct}
\end{figure*}

We therefore adopt a more challenging measurement structure for $\bA$ to better evaluate the algorithmic limits of BARM performance and reveal potential failures using both FoS \emph{and} FoRS metrics. Specifically, we first applied 2-D \emph{discrete cosine transform} (DCT) to $\bX_0$ and then randomly sampled $p$ of the resulting DCT coefficients. Because both the DCT and the sampling sub-process are linear operations on the entries of $\bX_0$, the whole process is representable via a matrix $\bA$, which encodes highly structured information.  Figure \ref{fig:dct} depicts the results using problem sizes consistent with Figure \ref{fig:amrm}; note that the FoRS metric has replaced the REL metric for comparison purposes.

Two things stand out from the analysis.  First, while the other algorithms display almost identical behavior under either metric, BARM failures under the FoS criteria are mostly converted to successes by the FoRS metric by recovering a matrix of near-optimal rank.  Secondly, even with this structured DCT-based sampling matrix, BARM outperforms the other algorithms using either metric.

\begin{table}[t]
\caption{Further matrix completion comparisons of BARM with IRLS0 by reducing the number of measurements in the hardest problem from \cite{mohan2012iterative}.  Results with both FoS and FoRS metrics are reported (avg of 10 trials).}
\label{tab:jmlr2}
\begin{center}
\begin{tabular}{ccc|cc|cc}
\hline
 \multicolumn{3}{c}{Problem} & \multicolumn{2}{c}{IRLSO} & \multicolumn{2}{c}{BARM} \\
\hline
 FR &  n(=m) & r  & FoS  & FoRS & FoS & FoRS \\
\hline

0.9  & 100  & 14 & 0 & 0  & 1 & 1\\
0.95  & 100 &  14 & 0 & 0 & 0.8 & 1\\
0.99  & 100 & 14 & 0 & 0 & 0.7 & 1\\
\hline
\end{tabular}
\end{center}
\end{table}

To summarize, we have demonstrated that BARM is capable of recovering a low-rank matrix right up to the theoretical limit in a variety of scenarios using different types of measurement processes.  Moreover, even in cases where it fails, it often nonetheless still produces a feasible $\hat{\bX}$ with rank nearly identical to the generative low-rank $\bX_0$, suggesting that multiple optimal solutions may be possible in challenging borderline cases.  But when true unequivocal failures do occur, such failures tend to be near the theoretical boundary, and with greater likelihood when the dictionary displays significant structure (or correlations).  While certainly we envision that, out of the infinite multitude of testing situations further significant pockets of BARM failure can be revealed, we nonetheless feel that BARM is quite promising relative to existing algorithms.

%\textbf{\emph{Application Examples:}}
%Although our primary purpose has been to derive and rigorously analyze BARM, and Monte-Carlo experiments with ground-truth data are arguably the most reliable way to do this, we have also conducted application-specific tests.  In the supplementary file we consider two such examples: image rectification and collaborative filtering for recommender systems.  The former implicitly involves a general sampling operator $\bA$, while the latter reduces to a standard matrix completion problem.

%Many real-world problems from disparate fields can be formulated as the search for a low-rank matrix under affine constraints  \cite{candes2009exact,candes2011robust,liu2013robust,zhang2012tilt,leger2010efficient}.

\subsection{Additional Noisy Tests}

\begin{figure}
	\centering
        \includegraphics[width = \columnwidth]{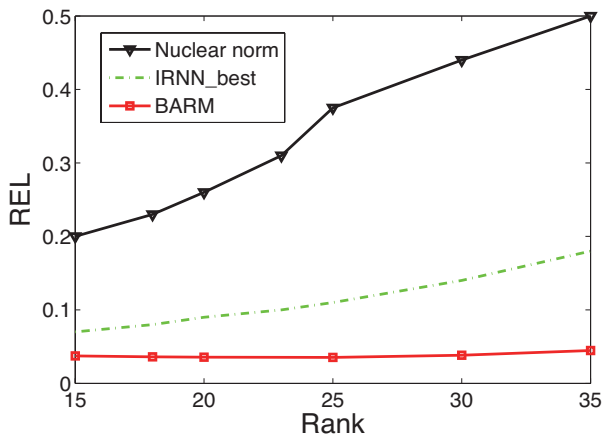}
    \caption{\small{Test with noisy data.}}
\label{fig:noisy}
\end{figure}

We also briefly present results that demonstrate the robustness of BARM to noise.  For this purpose we reproduce the noisy experiment from \cite{lu2014generalized} designed for validating IRNN algorithms.  The simulated data are generated in the exact same way as was used to produce Figure \ref{fig:mc}, only now instead of observing elements of $\bX_0$ directly, we observe $\bX_0 + 0.1\times\bE$, where elements of $\bE$ are iid $\calN(0,1)$.  Although in \cite{lu2014generalized} a heuristic strategy is introduced and tuned for adaptively setting all parameters (four in total), we simply applied BARM with $\lambda = 10^{-3}$ (so only a single parameter need be adjusted, and actually a wide range of $\lambda$ values produces similar performance anyway).  Results are shown in Figure \ref{fig:noisy} where we compare BARM directly with the best result reported in \cite{lu2014generalized} over the range $r = 15$ to $r=35$.  The nuclear norm solution is also included for reference.  Overall, the BARM solution is stable and exhibits superior accuracy relative to the others.

%In the presence of noise, again we choose to reproduce the exact same experiment from \cite{lu2014generalized}. Specifically, a rank $r$ matrix is first generated as $\bX_0 = \mathbf{M}_L \mathbf{M}_R$, with $\mathbf{M}_L \in \mathbb{R}^{n \times r}$  and $\mathbf{M}_R \in \mathbb{R}^{r \times m}$ ($n=m=150$) as iid $\calN(0,1)$ random matrices. Then a noisy observation $\bY$ is generated as $\bY = \bX_0+0.1\times \bE$, where $\bE$ is an iid $\calN(0,1)$ random noise matrix. 50\% of all entries of $\bY$ are then hidden uniformly at random.  Although in \cite{lu2014generalized}, the authors have introduced a complex strategy for adaptively setting all parameters, we simply applied BARM with $\lambda_1=0.1$ and $\lambda_2=10^{-6}$ for all the tests. Results are shown in Figure \ref{fig:noisy}, where we compare the performance of BARM directly with the best results  reported in \cite{lu2014generalized}.

\subsection{Comparisons with Rank-Aware Algorithms}
\label{ssec:com2alter}

As stated previously, a somewhat different class of non-convex algorithms were derived by taking advantage of the correct rank and alternatively updating certain rank-dependent decompositions of the original matrix, e.g. \cite{jain2013low}, \cite{tanner2013normalized}.  While given the correct rank, these algorithms (and their generalizations) were reported to have achieved performance relatively near to theoretical limits in special circumstances \cite{tanner2013normalized,tanner2014low}. Here we delve into comparisons of BARM with these algorithms.

\begin{figure*}
	\centering
    \subfigure[$50 \times 50$, $\bA$ uncorrelated]
    {
        \label{fig:amrm50noa}
        \includegraphics[width = 0.96\columnwidth]{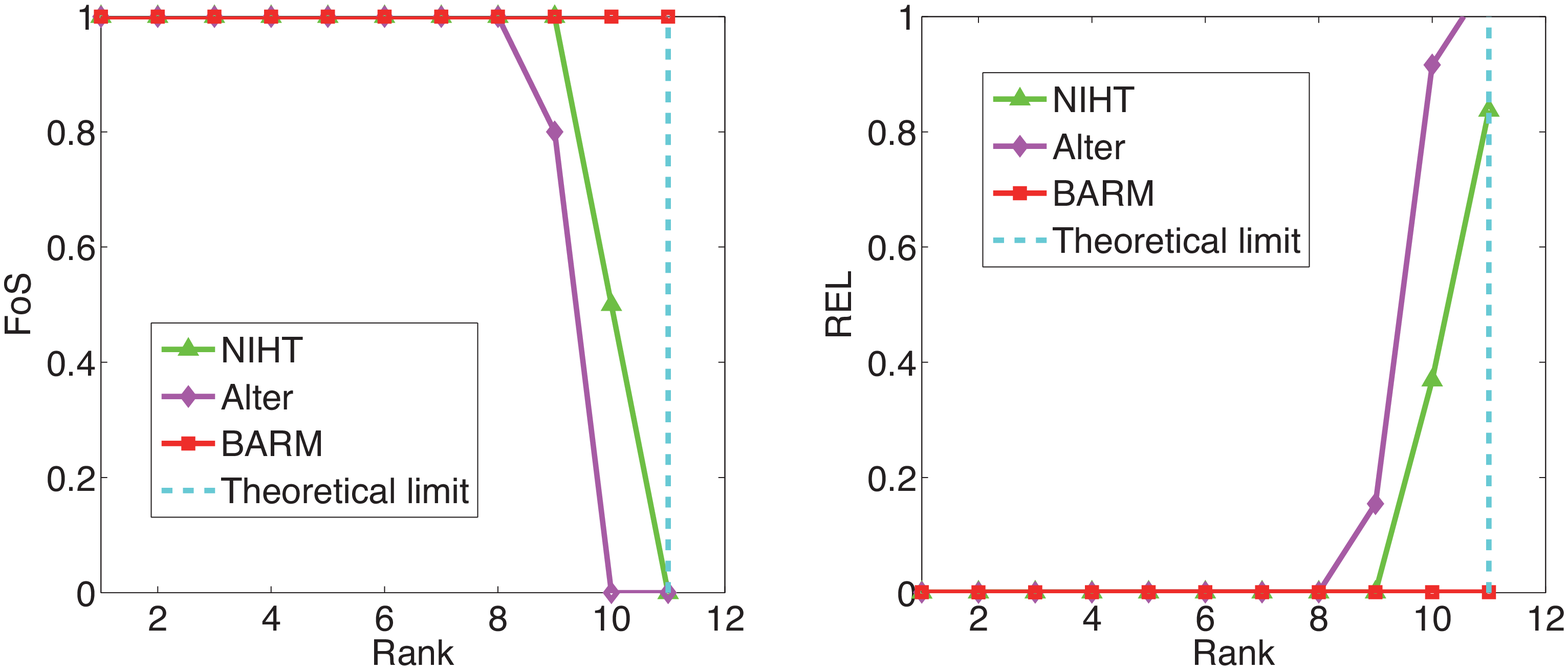}
    }
    \subfigure[$50 \times 50$, $\bA$ correlated]
    {
        \label{fig:amrm50yesa}
        \includegraphics[width = 0.96\columnwidth]{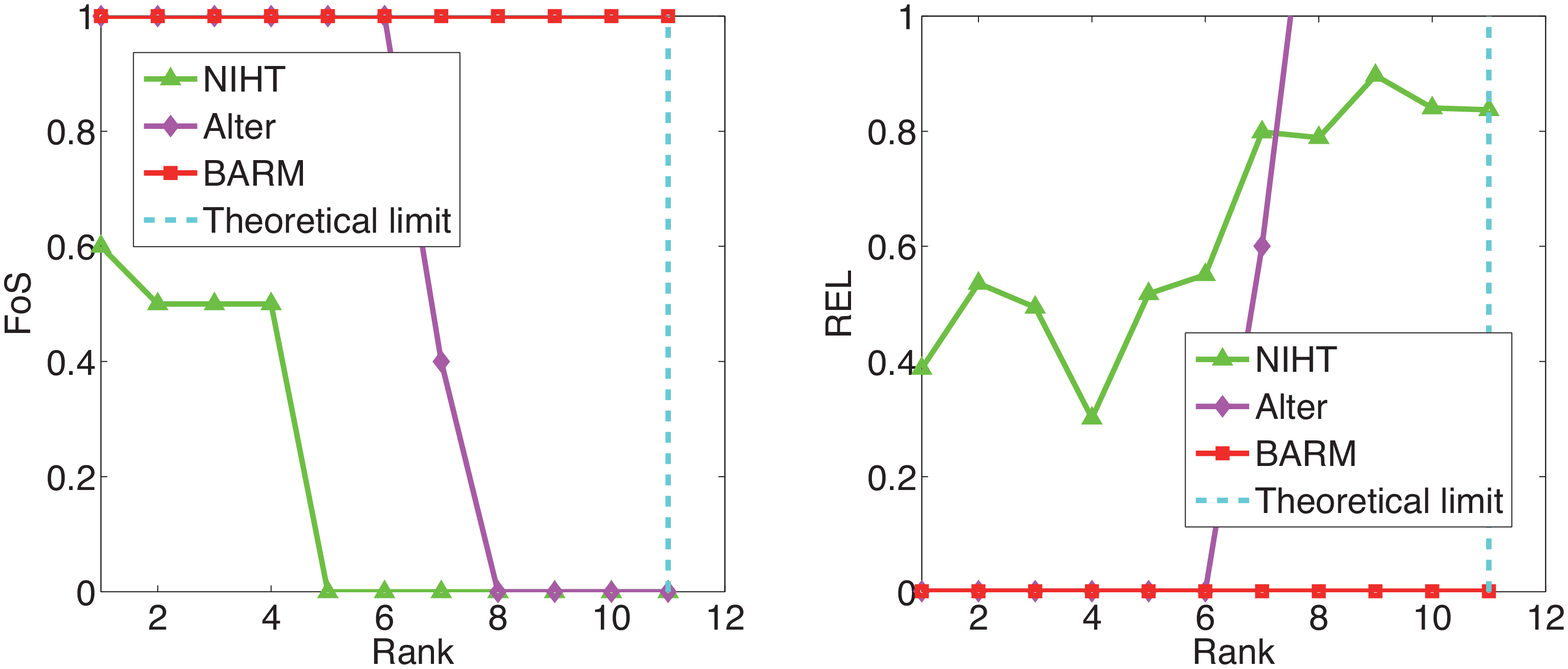}
    }
    \vspace{-2mm}
    \subfigure[$100 \times 100$, $\bA$ uncorrelated]
    {
        \label{fig:amrm100noa}
        \includegraphics[width = 0.96\columnwidth]{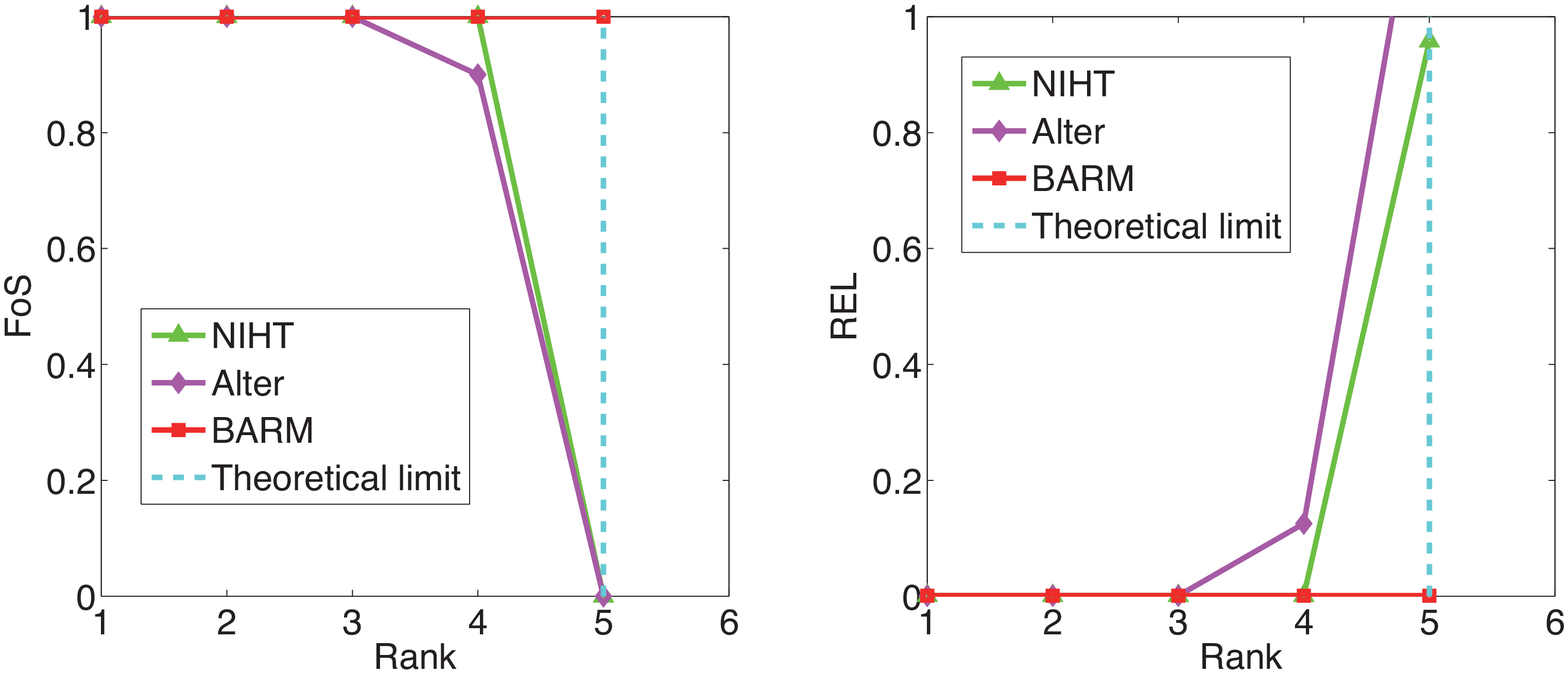}
    }
    \subfigure[$100 \times 100$, $\bA$ correlated]
    {
        \label{fig:amrm100yesa}
        \includegraphics[width = 0.96\columnwidth]{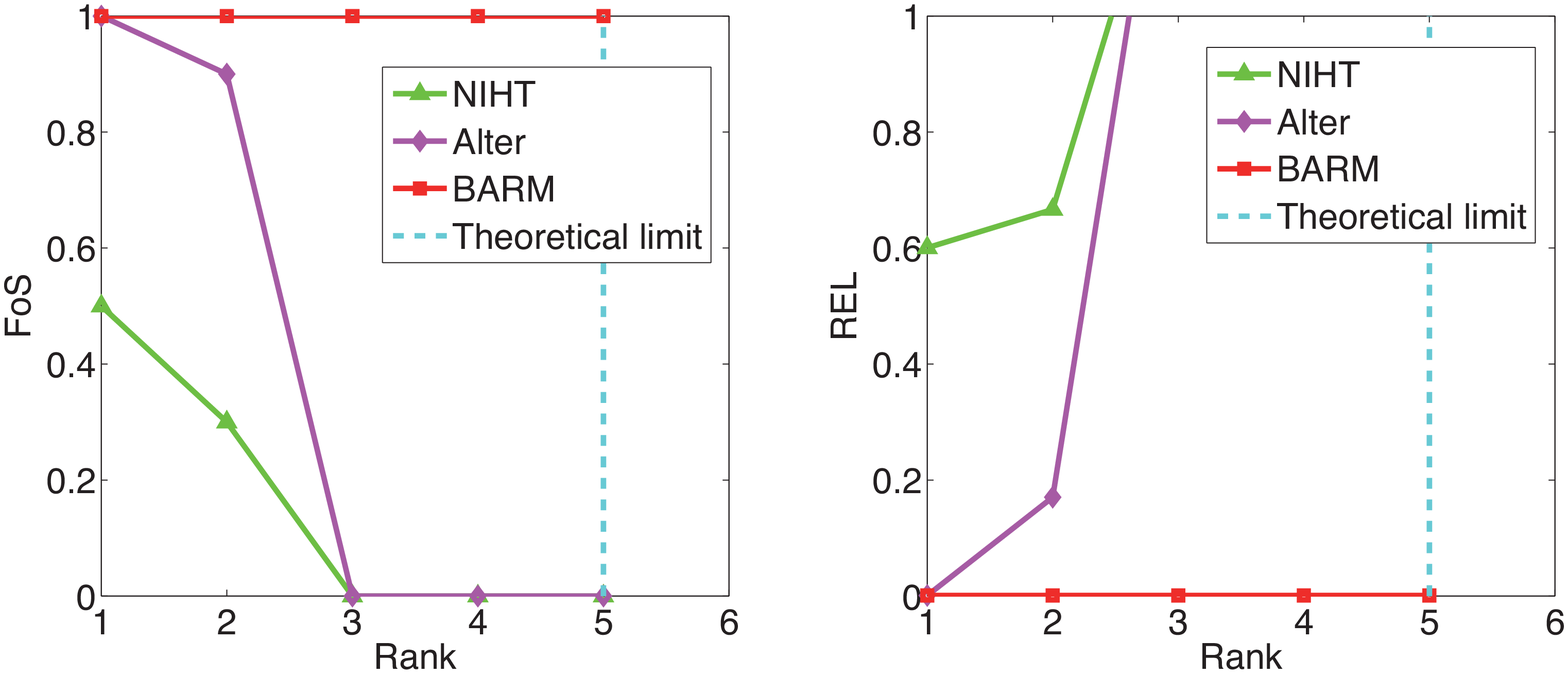}
    }
	\caption{\small{Results here reproduce comparisons of Figure \ref{fig:amrm}, but with rank-aware algorithms NIHT and Alter.  BARM has no knowledge of the true rank.}}
\label{fig:amrma}
\end{figure*}

\begin{figure}
    \centering
    \includegraphics[width = 0.95\columnwidth]{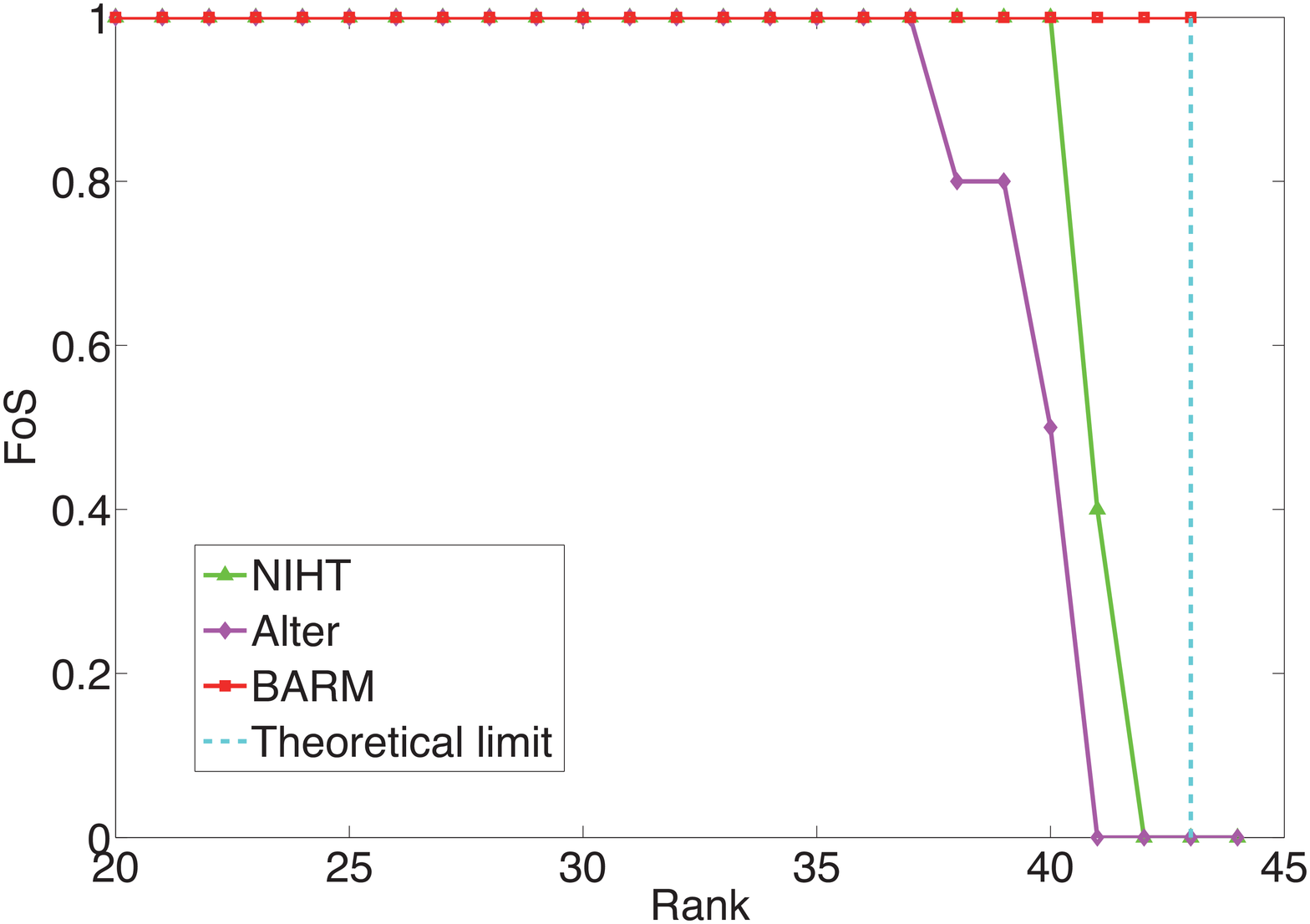}
    \caption{\small{Results here reproduce the comparisons of Figure \ref{fig:mc}, but with rank-aware algorithms NIHT and Alter.  BARM has no knowledge of the true rank.}}
    \label{fig:mc2}
\end{figure}

\begin{figure}
    \centering
    \includegraphics[width = 0.95\columnwidth]{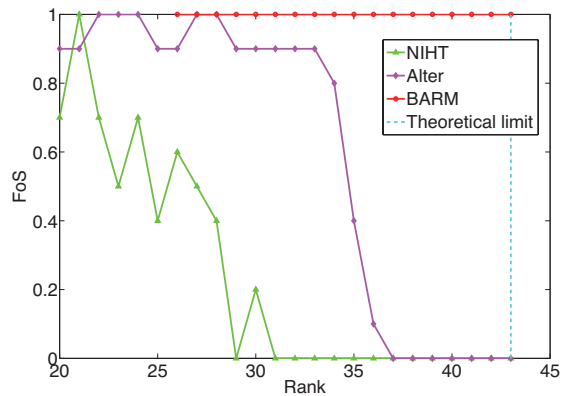}
    \caption{\small{Matrix completion comparisons with $\bX_0$ having decaying singular values (avg of 10 trials)}}
    \label{fig:mc3}
\end{figure}

Specifically, we denote the algorithms in \cite{jain2013low}, \cite{tanner2013normalized} as Alter (short for alternating algorithm) and NIHT (following \cite{tanner2013normalized}) respectively. We tested both on the tasks introduced in Section \ref{ssec:mc} and \ref{ssec:gen}. Figure \ref{fig:amrma} and Figure \ref{fig:mc2} illustrate the corresponding results. Given the correct rank a priori, Alter and NIHT generally achieve better results compared to IRLS0 and the nuclear norm.  However, even with the correct rank known, their performance cannot match BARM, especially so on more challenging tasks as shown in Figure \ref{fig:amrm50yesa} and \ref{fig:amrm100yesa}. Moreover, this type of algorithm can perform worse when $\bX_0$ has decaying singular values. We empirical observe such phenomenon in Figure \ref{fig:mc3}. In this case, we first generate $\bX_0$ as above and then multiply its $i$th (largest) singular value by a factor of $\frac{1}{i^{0.8}}$, leading to a new $\bX_0$ with moderately decaying singular values. Note that, while the performance of both Alter and NIHT dropped evidently, BARM still achieved 100\% success till the limit.

\subsection{Computational Complexity} \label{sec:complexity}

Finally, regarding computational complexity, for general $\bA$ the BARM updates can be implemented to scale linearly in the elements of $\bX$ and quadratically in the number of observations $p$ (the special case of matrix completion is decidedly much cheaper because of the special structure that can be exploited).  In our experiments, for relatively easy problems on the order of 10 iterations are required, while for difficult recovery problems near the theoretical recovery boundary this may increase by a factor of 10 or so.  This is somewhat expected though since as we near the theoretical limit, $\bA$ becomes highly overcomplete, and candidate solutions become much more difficult to differentiate.

\begin{figure}
	\centering
\includegraphics[width = \columnwidth]{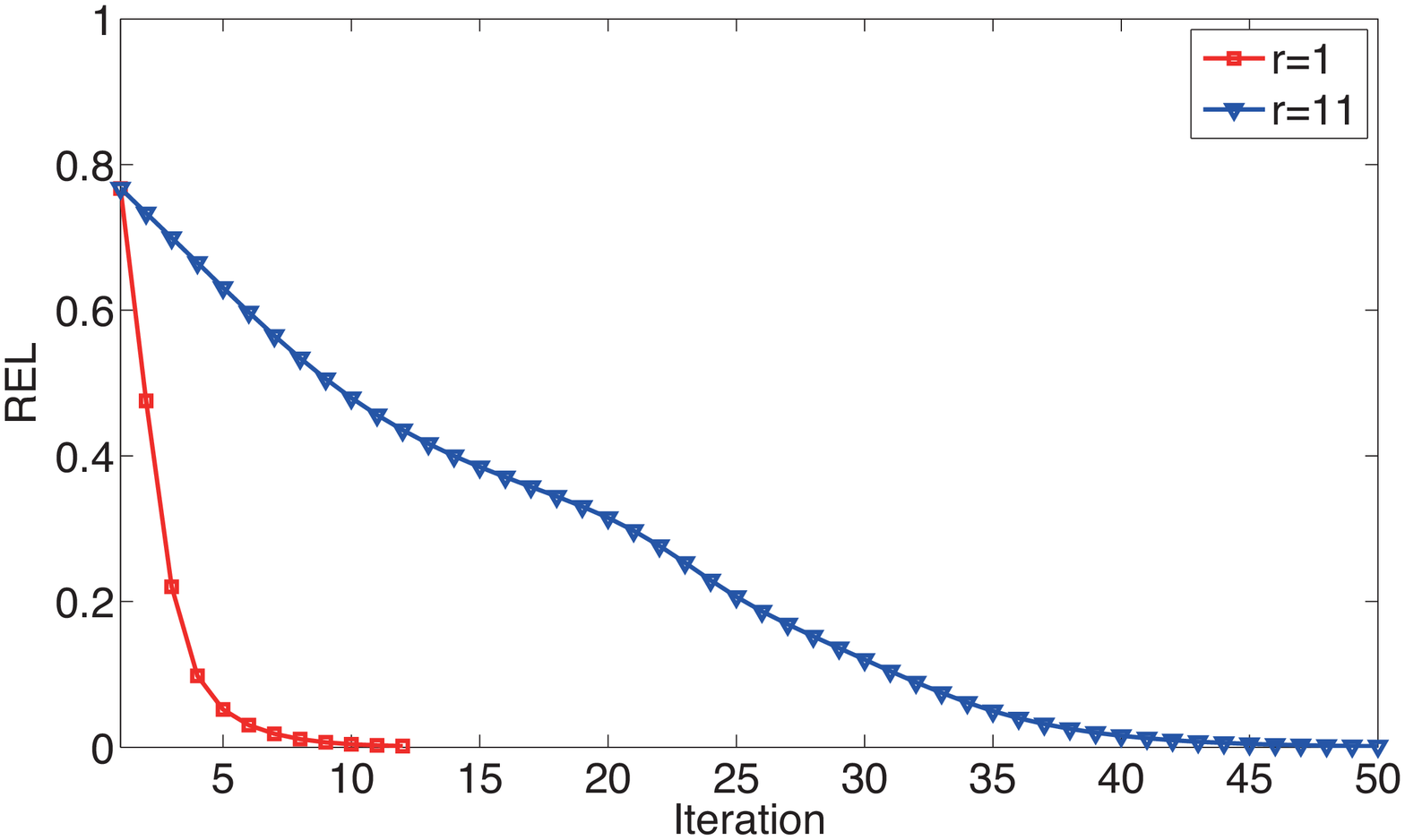}
    \caption{\small{Empirical convergence of BARM.}}
\label{fig:conv}
\end{figure}

To show this effect empirically, we compare two separate trials from Figure \ref{fig:amrm}(a), the first when $r=1$ (relatively easy), the second when $r=11$ (relatively hard).\footnote{Note that $r=1$ is only relatively easy here because the number of observations is sufficient for the larger $r=11$ case; if only the minimal number of measurements are available then even $r=1$ can be challenging for many algorithms.}  In Figure \ref{fig:conv} we plot the value of REL in both cases versus the iteration number of BARM.

\section{Application Examples}
\label{sec:app}

Many real-world problems from disparate fields can be formulated as the search for a low-rank matrix under affine constraints   \cite{candes2009exact,liu2013robust,zhang2012tilt,leger2010efficient}.  Here we briefly consider two such examples: low-rank image rectification and collaborative filtering for recommender systems.  The former implicitly involves a general sampling operator $\bA$, while the latter reduces to a standard matrix completion problem.

%\textbf{\emph{Low-rank Image Rectification:}}
\subsection{Low-rank Image Rectification} \label{sec:image_rectification}

In \cite{zhang2012tilt}, the \emph{transform invariant low-rank textures} (TILT) algorithm is derived for rectifying images containing low-rank textures that have been transformed using an unknown operator $\btau$ from some group (e.g., a homography).  For a given observed image $\bY$, the basic idea is to construct a first-order Taylor series approximation around the current rectified image estimate $\hat{\bX}$ and solve
\begin{equation}
\label{eq:tilt}
    \min_{\bX,\bdelta} ~~ \rank[\bX]~~~\mbox{s.t.}~\bX = \bY + \sum_{i} \bJ_i(\hat{\bX}) \delta_i,
\end{equation}
where $\bJ_i(\hat{\bX})$ is the Jacobian matrix with respect to $\bX$ of the $i$-th parameter $\tau_i$ describing the transformation, with $\btau = [\tau_1,\tau_2,\ldots]^{\top}$.  Optimization over the vector of first-order differences $\bdelta = [\delta_1,\delta_2,\ldots]^{\top}$ can be accomplished in closed form by projecting both sides of the constraint to the orthogonal complement of the span of all $\bJ_i(\hat{\bX})$.  Let $P_{\bJ^c}$ represent this projection operator.  The feasible region in (\ref{eq:tilt}) then becomes
\begin{equation}
P_{\bJ^c}\left( \bX \right) = P_{\bJ^c}\left( \bY \right) + P_{\bJ^c}\left( \sum_{i} \bJ_i(\hat{\bX}) \delta_i \right) = P_{\bJ^c}\left( \bY \right)
\end{equation}
The resulting problem then reduces exactly to (\ref{eq:affine_rank_problem}) when we define $\mathcal{A} = P_{\bJ^c}$ and $\bb = \myvec \left[ P_{\bJ^c}\left( \bY \right) \right]$.  Once $\bX$ is computed in this way, we then update each $\bJ_i(\hat{\bX})$ and repeat until convergence.

\begin{figure*}
	\centering
    \subfigure
    {
        \includegraphics[width = 0.46\columnwidth]{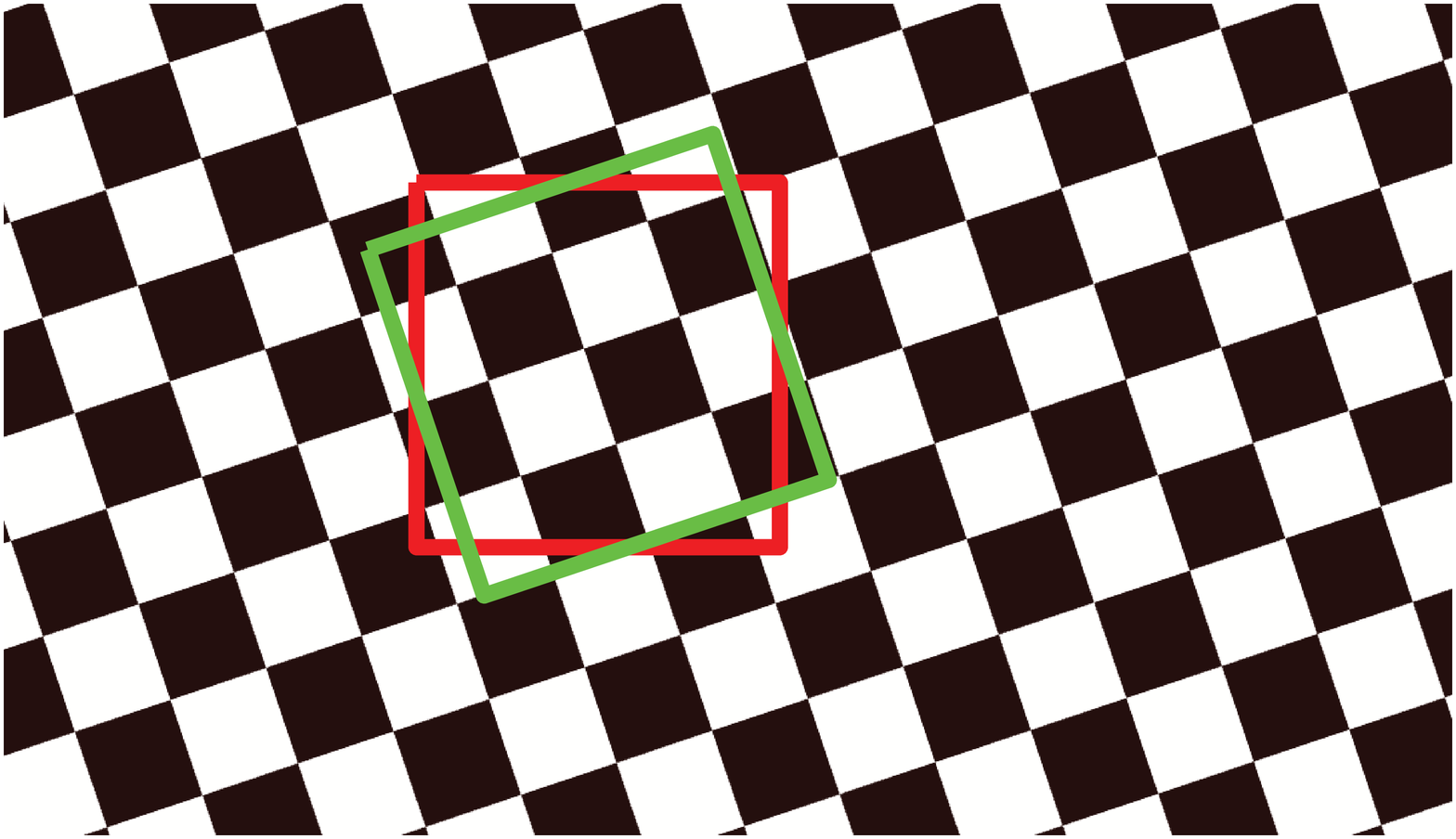}
    }
    \subfigure
    {
        \includegraphics[width = 0.46\columnwidth]{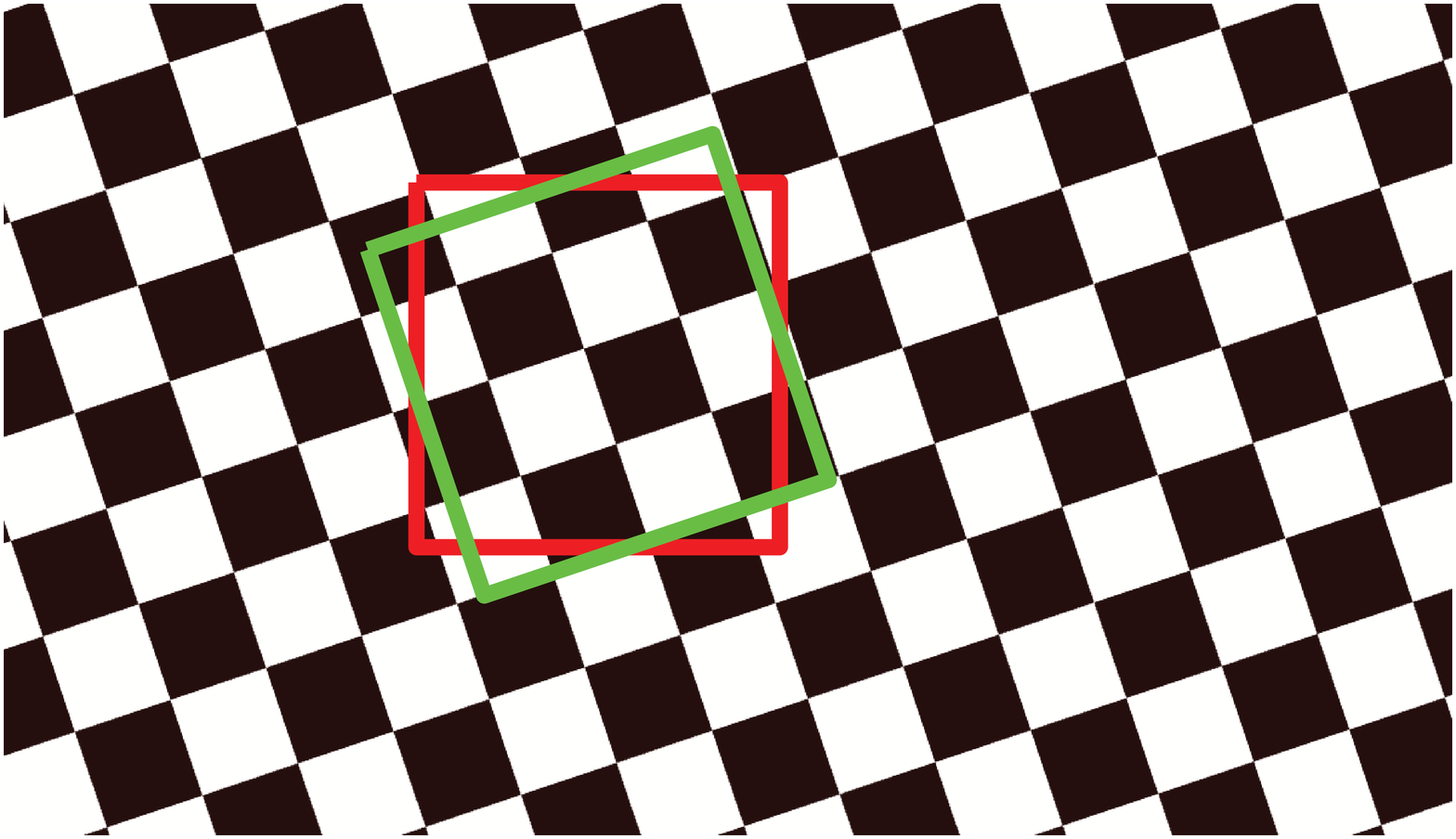}
    }
    \subfigure
    {
        \includegraphics[width = 0.46\columnwidth]{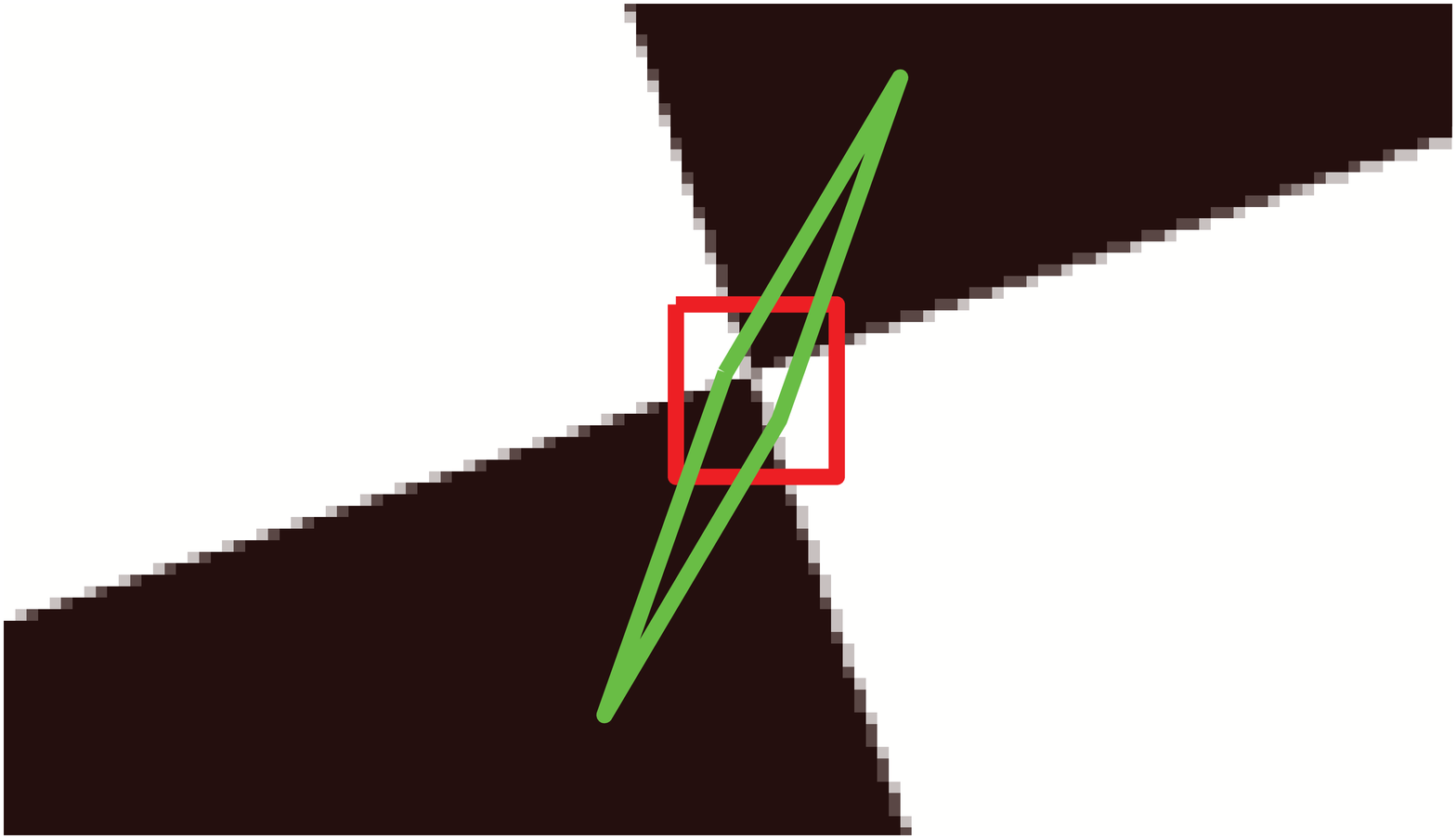}
    }
    \subfigure
    {
        \includegraphics[width = 0.46\columnwidth]{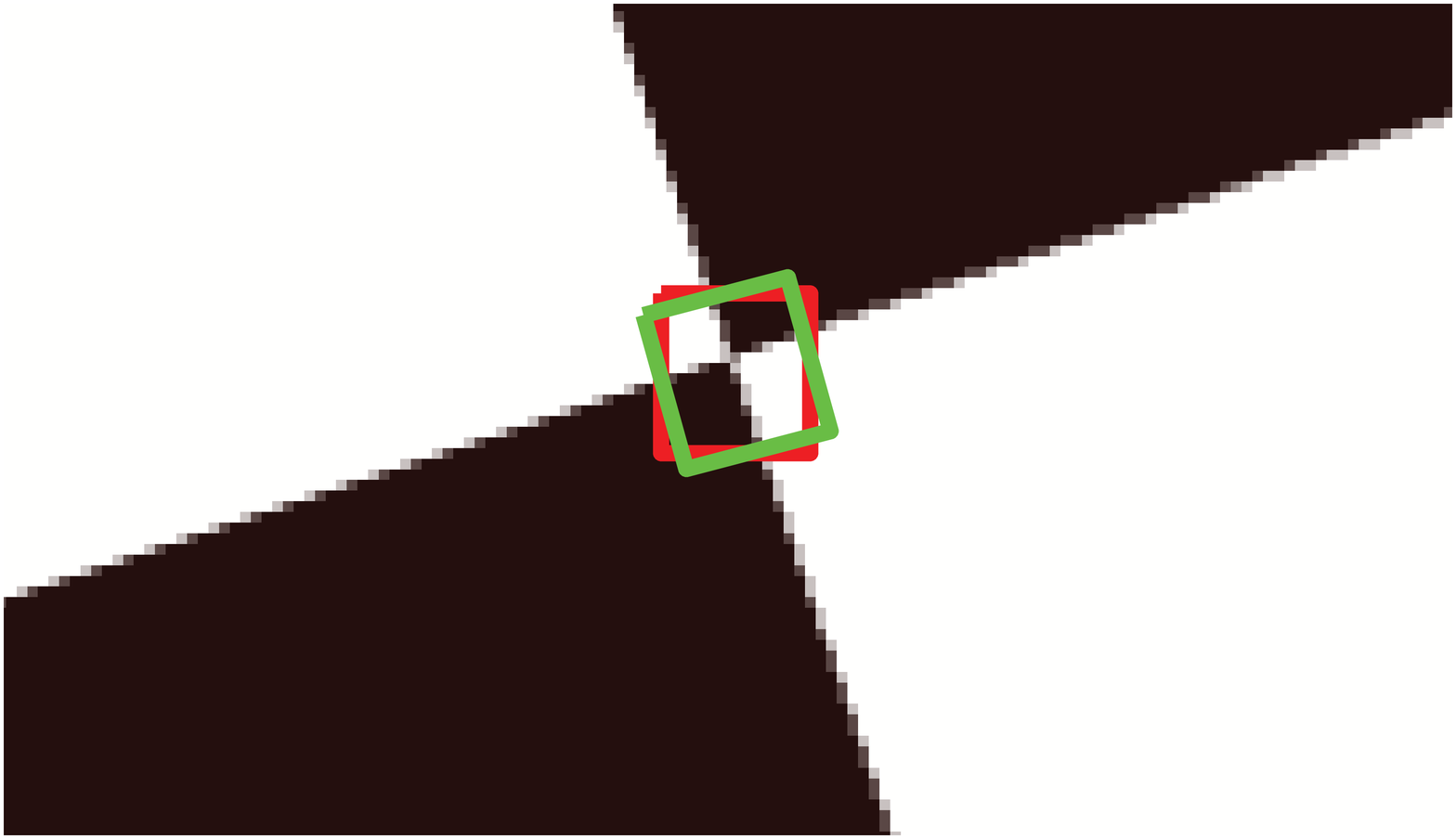}
    }
    \vspace{-3mm}
     \setcounter{subfigure}{0}
    \subfigure[Nuclear norm (easy)]
    {
        \includegraphics[width = 0.46\columnwidth]{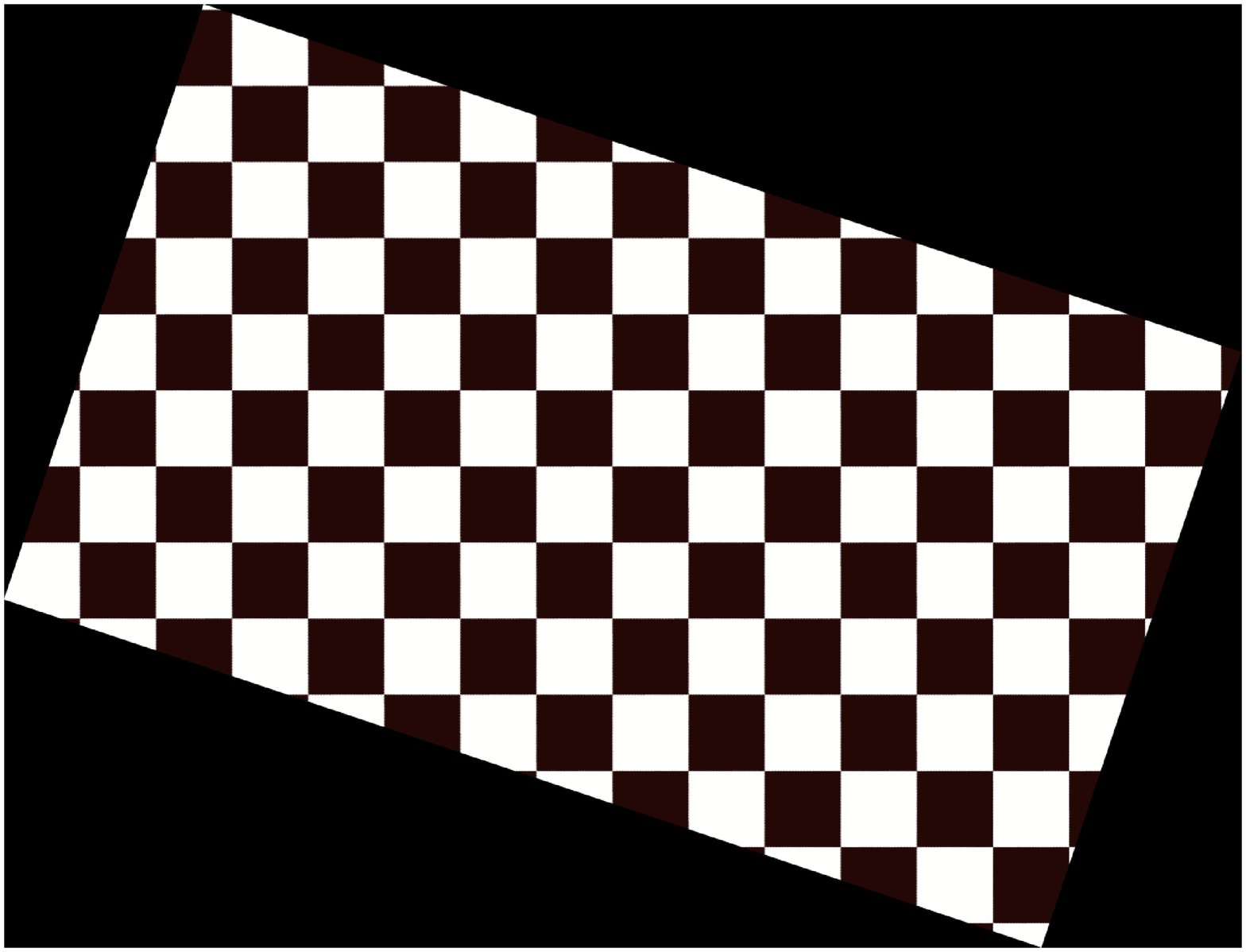}
    }
    \subfigure[BARM (easy)]
    {
        \includegraphics[width = 0.46\columnwidth]{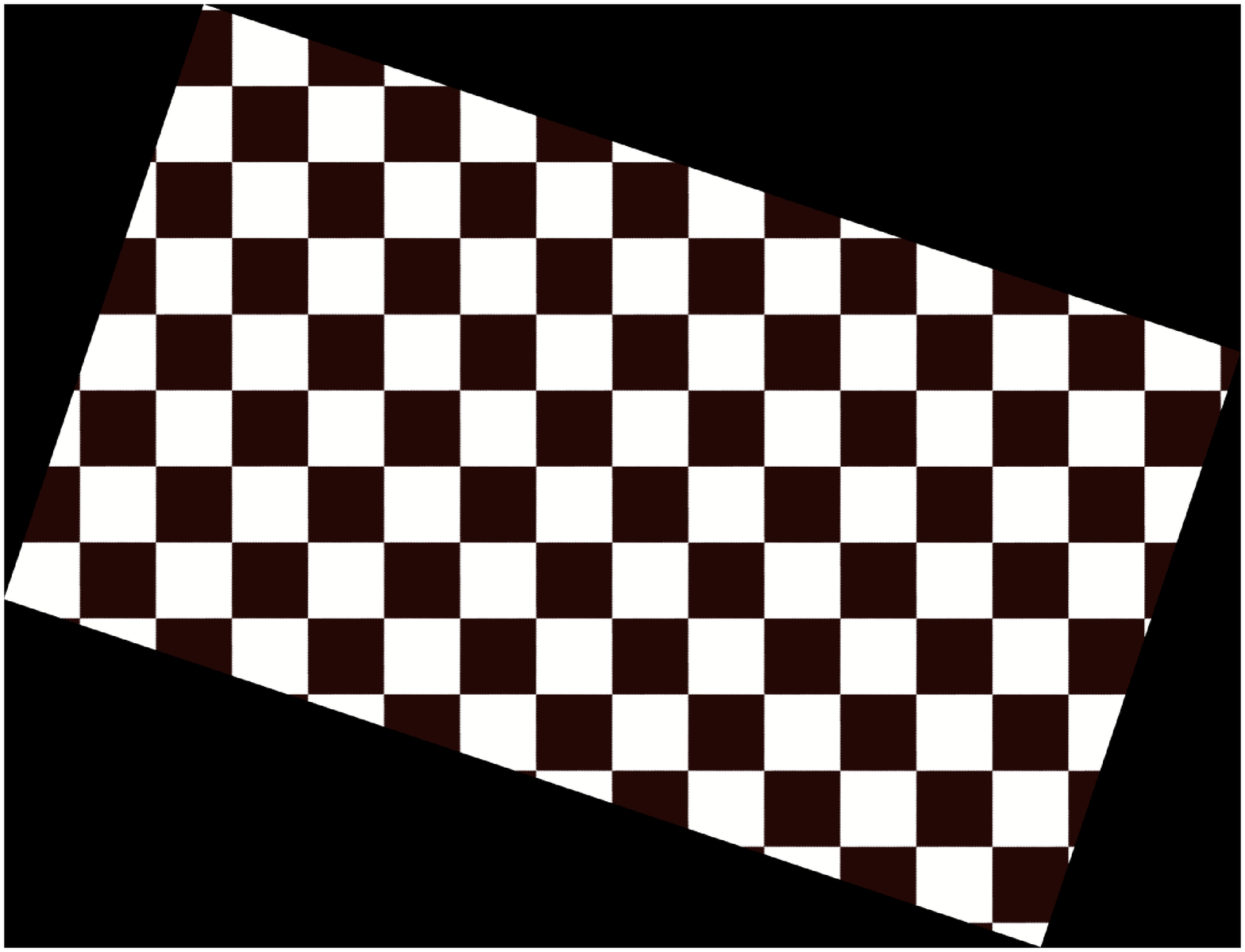}
    }
    \subfigure[Nuclear norm (hard)]
    {
        \includegraphics[width = 0.46\columnwidth]{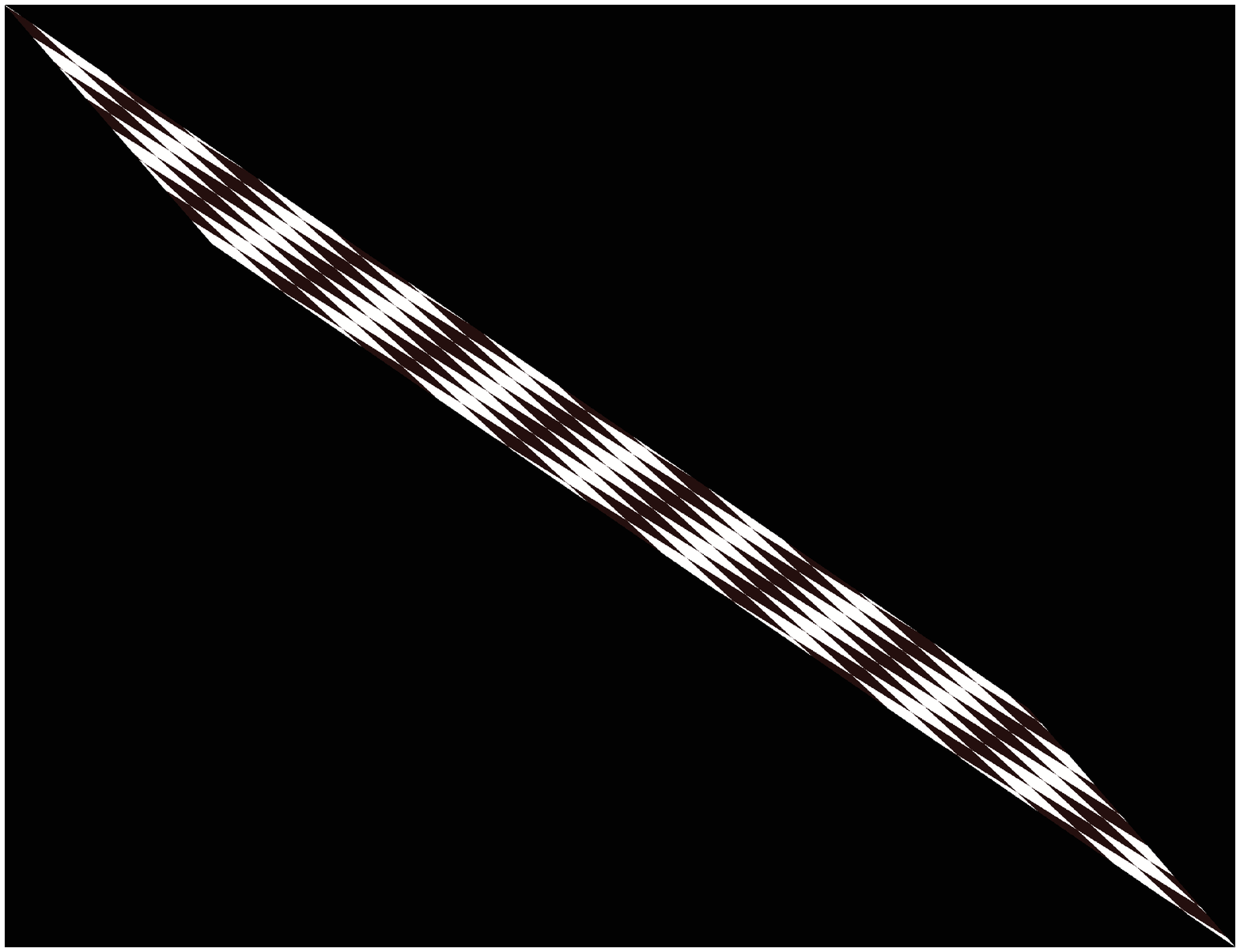}
    }
    \subfigure[BARM (hard)]
    {
        \includegraphics[width = 0.46\columnwidth]{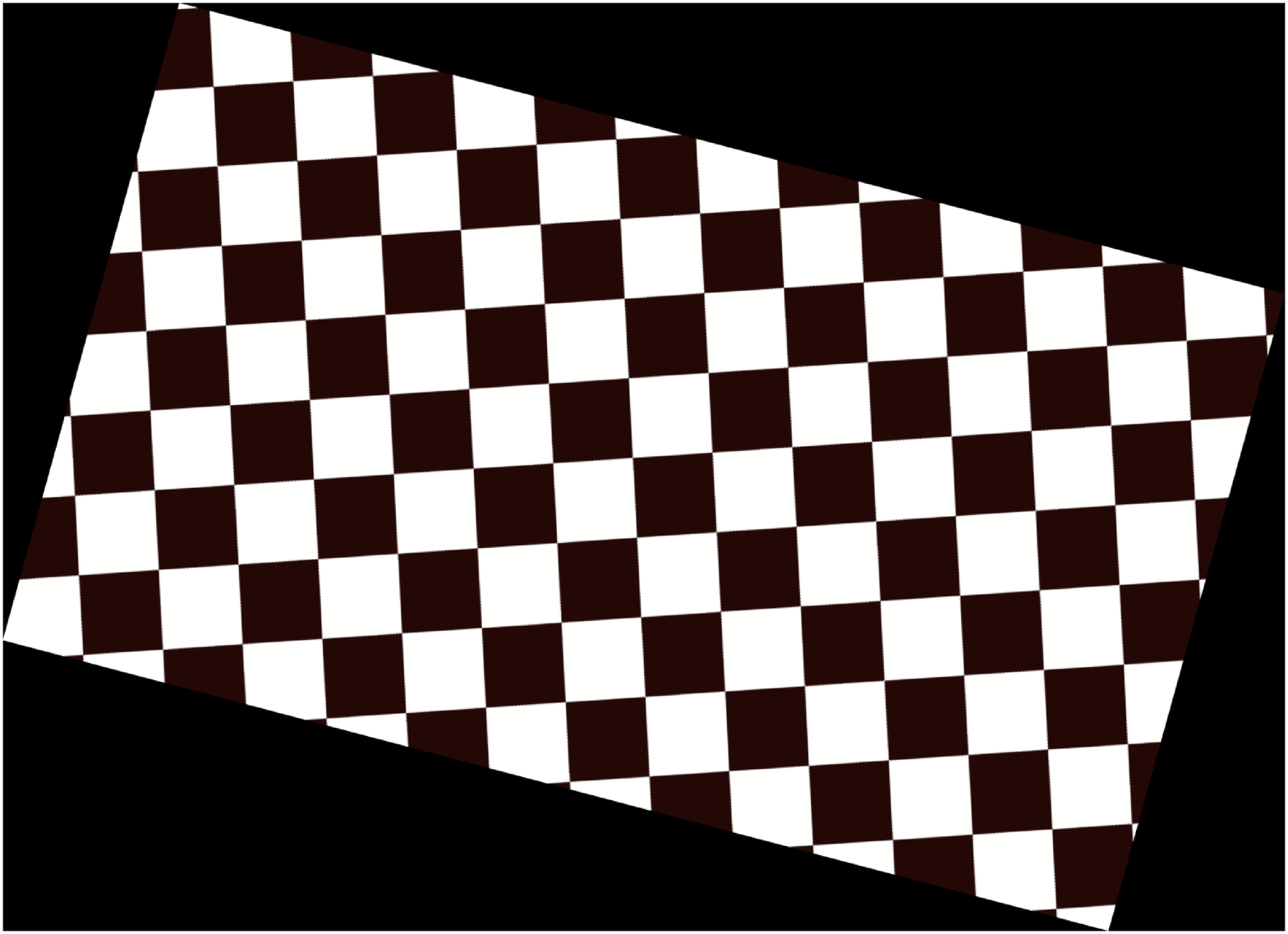}
    }
	\caption{\small{Image rectification comparisons using a checkboard image. \emph{Top}: Original image with observed region (red box) and estimated transformation (green box). \emph{Bottom}: Rectified image estimates.}}
\label{fig:tilt1}
\end{figure*}

\begin{figure*}
	\centering
    \subfigure
    {
        \includegraphics[width = 0.46\columnwidth]{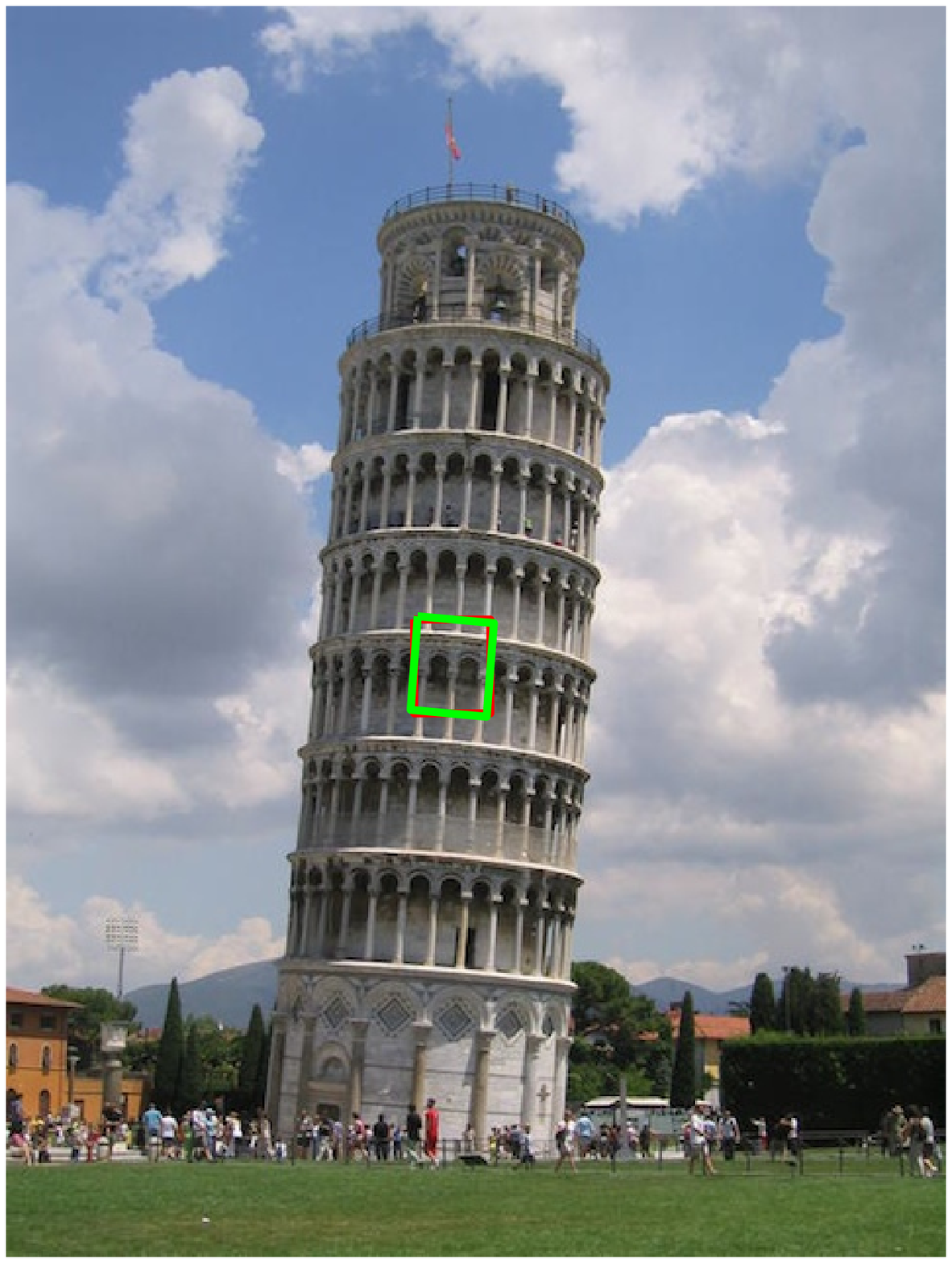}
    }
    \subfigure
    {
        \includegraphics[width = 0.46\columnwidth]{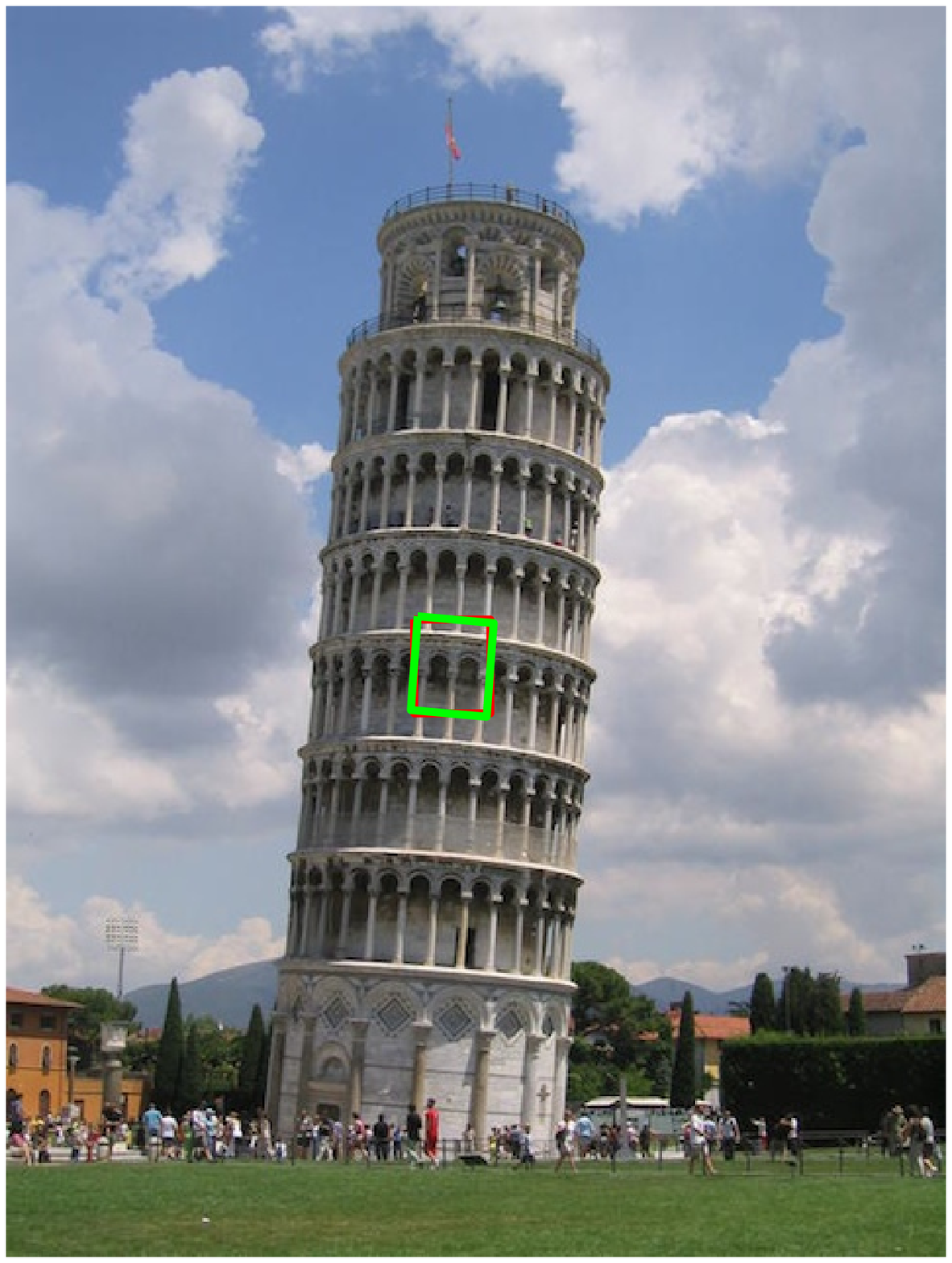}
    }
    \subfigure
    {
        \includegraphics[width = 0.46\columnwidth]{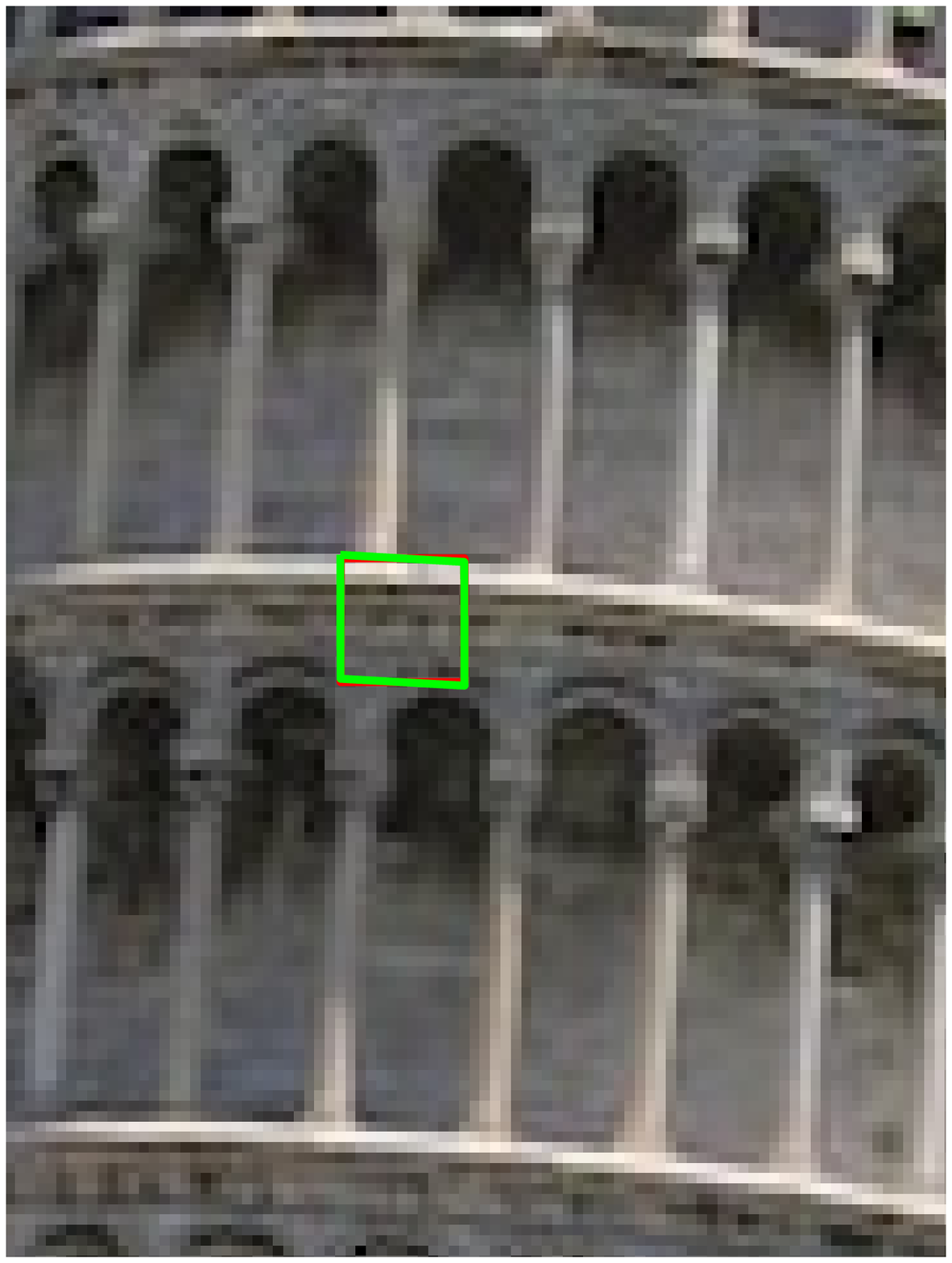}
    }
    \subfigure
    {
        \includegraphics[width = 0.46\columnwidth]{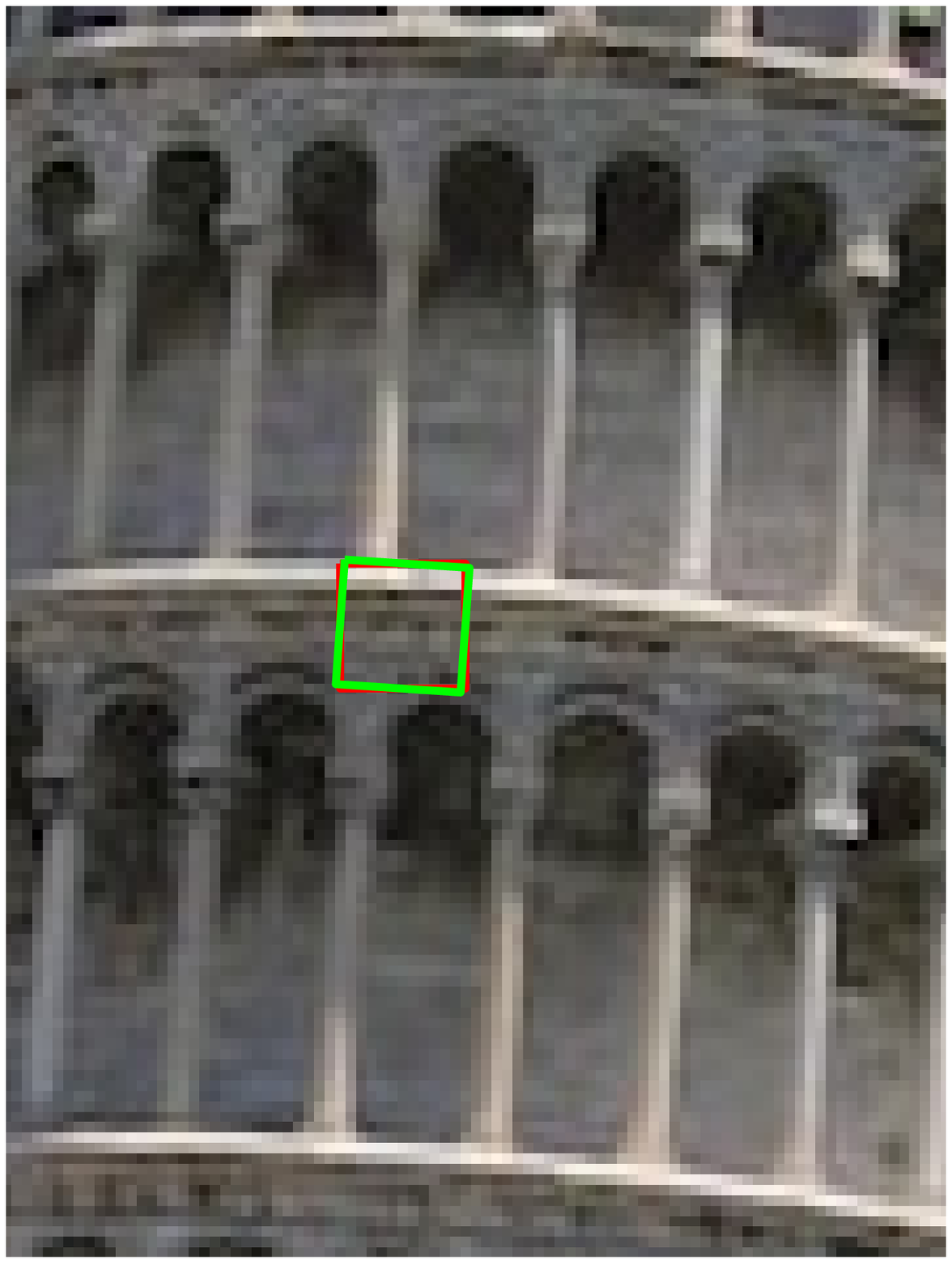}
    }
    \vspace{-3mm}
     \setcounter{subfigure}{0}
    \subfigure[Nuclear norm (easy)]
    {
        \includegraphics[width = 0.46\columnwidth]{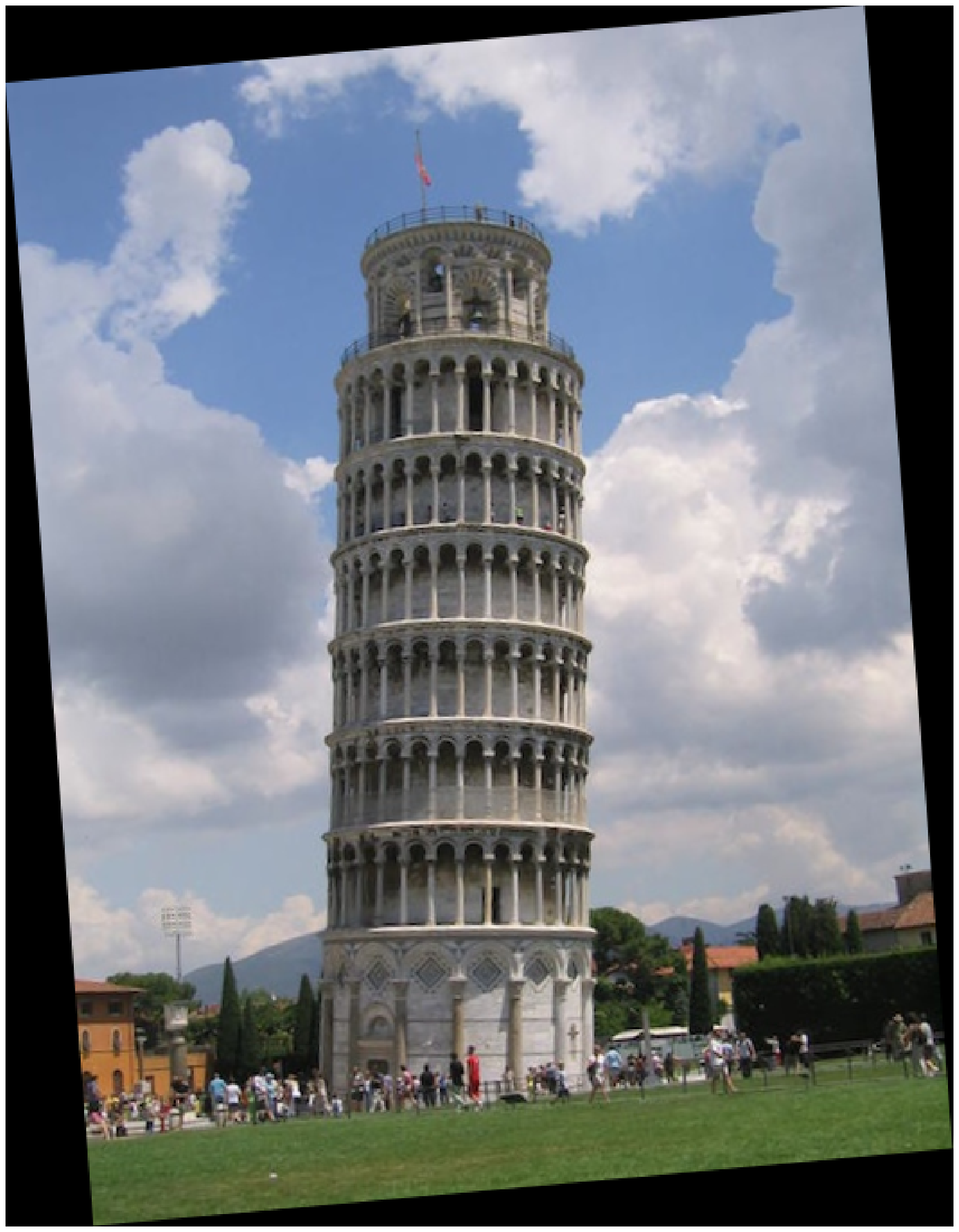}
    }
    \subfigure[BARM (easy)]
    {
        \includegraphics[width = 0.46\columnwidth]{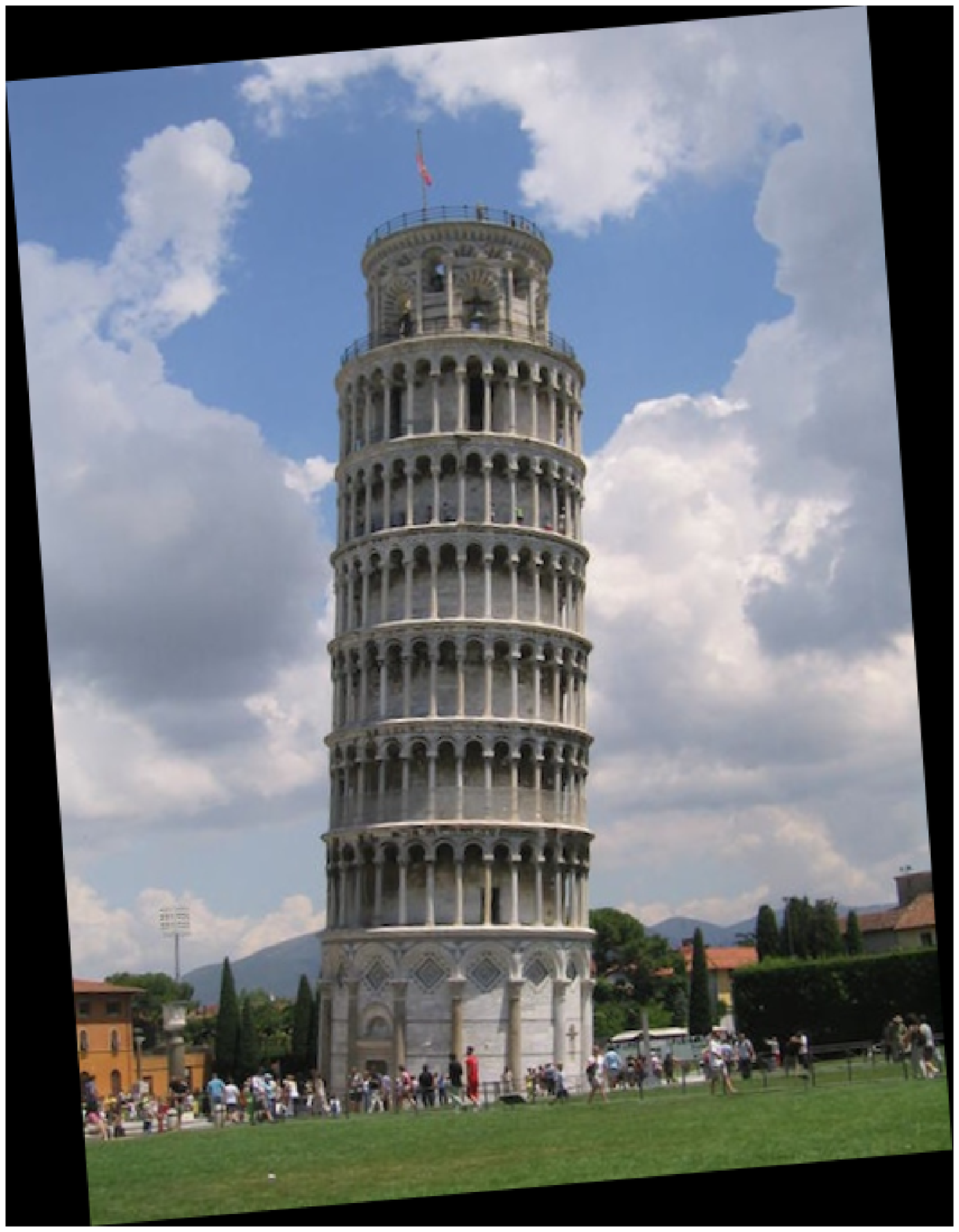}
    }
    \subfigure[Nuclear norm (hard)]
    {
        \includegraphics[width = 0.46\columnwidth]{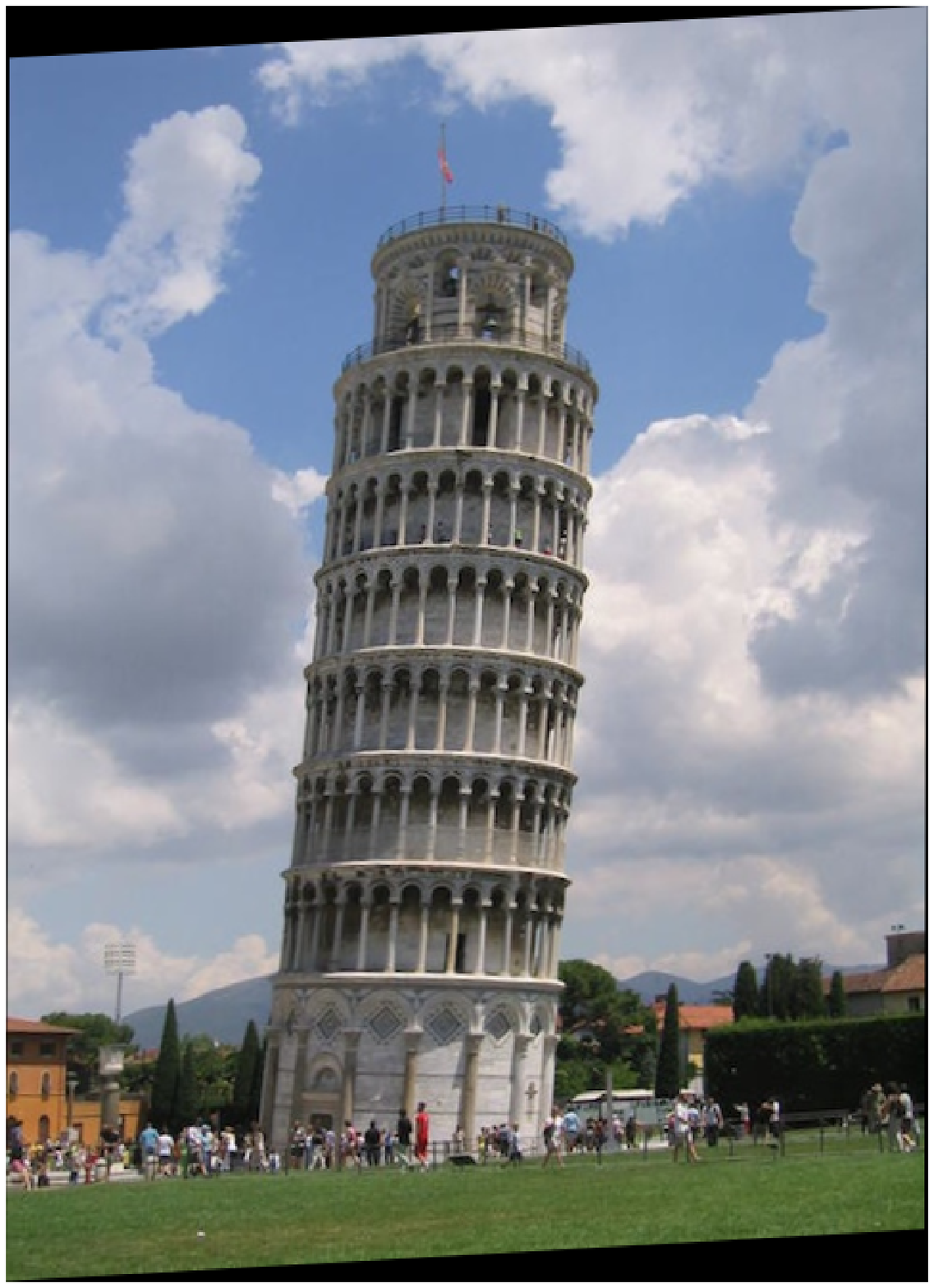}
    }
    \subfigure[BARM (hard)]
    {
        \includegraphics[width = 0.46\columnwidth]{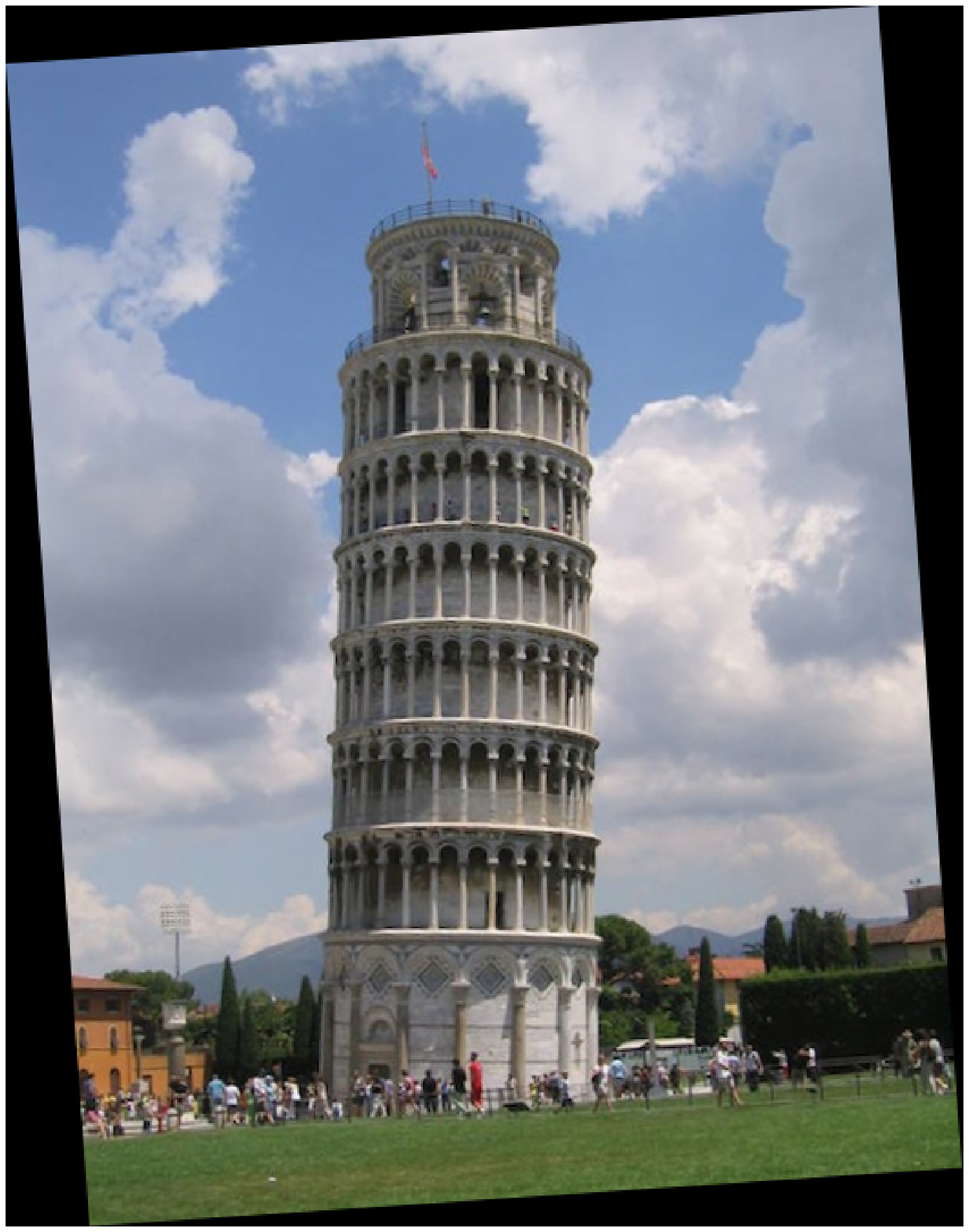}
    }
	\caption{\small{Image rectification comparisons using a landmark photo. \emph{Top}: Original image with observed region (red box) and estimated transformation (green box). \emph{Bottom}: Rectified image estimates.}}
\label{fig:tilt2}
\end{figure*}

While the original TILT algorithm substitutes the nuclear norm for $\rank[\bX]$, we embedded the BARM algorithm into the posted TILT source code \cite{zhang2012tilt} for comparison purposes (note that we disabled an additional sparse error term for both algorithms to simplify comparisons, and it is not necessary anyway in many regimes).  Figures \ref{fig:tilt1} and \ref{fig:tilt2} display results on both two easy examples, where the number of observations $p$ is large, and two more difficult problems where the number observations is small.  While both algorithms succeed on the easy cases, when the observations are constrained by a small image window, only BARM is successful in accurately rectifying the images.  This may be due, at least in part, to the fact that the implicit $\mathcal{A}$ operator contains significant structure that is not consistent with the required nullspace properties required for nuclear norm minimization success.

% \textbf{\emph{MovieLens Matrix Completion:}}
\subsection{Collaborative Filtering of MovieLens Data}

Collaborative filtering, a technique used by many recommender systems, is a popular representative application of low-rank matrix completion. Typically the rows (or columns) of $\bX_0$ index users, the columns (or rows) denote items, and each entry $(\bX_0)_{ij}$ is the rating/score of user $i$ applied to item $j$. Given that we can observe some subset of elements of $\bX_0$, the task of collaborative filtering is to predict all or some of the missing ratings.  In general this would be impossible; however, if we have access to some prior knowledge, e.g., $\bX_0$ is low-rank, then estimation may be feasible.

While our interest here is not in recommender systems or collaborative filtering per se, we nonetheless evaluate BARM using the 1M MovieLens dataset\footnote{http://www.grouplens.org/} as this appears to represent one of the most common evaluation benchmarks.  We emphasize at the outset that the strict validity of any low-rank assumptions underlying this data is debatable, and it remains entirely unclear whether the true globally optimal or lowest rank solution consistent with the observations, even if computable, would necessarily lead to the best prediction of the unknown ratings.  In fact, the reported performance of various existing rank-minimization algorithms tends to cluster around almost the same value, implying that collaborative filtering may not provide the most discriminative data type with which to compare.  In most cases, it appears that tuning parameters and other heuristic modifications play a larger role than the underlying algorithmic distinctions fundamental to finding optimal low-rank estimates.  Nonetheless, we apply BARM for completeness and convention, adopting an additional simple mean-offset estimation term from \cite{leger2010efficient} that is particularly suitable for this problem.

\begin{table}[t]
\caption{Collaborative filtering on 1M MovieLens dataset. Results from \cite{leger2010efficient} are in italic for comparison purposes.}
\label{tab:movie}
\begin{center}
\begin{tabular}{c|cc}
\hline
  & Weak NMAE & Hard NMAE \\
\hline
\it{URP}  & \it{0.4341} & \it{0.4444} \\
\it{Attitude} & \it{0.4320}  & \it{0.4375} \\
\it{MMMF} & \it{0.4156}  & \it{0.4203} \\
\it{IPCF} & \it{0.4096}  & \it{0.4113} \\
\it{E-MMMF}  & \it{0.4029} & \it{0.4071} \\
\it{GPLVM} & \it{0.4026}  & \it{0.3994} \\
\it{NBMC} & \bf{{0.3916}}  & \it{0.3992} \\
\it{IRLS/GM} & \it{0.3959}  & \it{0.3928} \\
\hline
%IRLS/GM(our) & {0.3945}  & {0.3900} \\
BARM & {0.3942}  & \bf{0.3898} \\
\hline
\end{tabular}
\end{center}
\end{table}

In \cite{mohan2012iterative}, IRLS0 is compared with only two other algorithms on MovieLens data, but the performance is no better.  Therefore, we choose to compare directly with \cite{leger2010efficient}, which both derives an IRLS-like algorithm and shows comparisons with a much wider variety of alternative algorithms using a strict evaluation protocol that is standard in the literature.  Specifically, the 1M MovieLens dataset, which contains 1 million ratings in the range $\{1,...,5\}$ for 3900 movies from 6040 unique users, is assessed under two test-protocals: \emph{weak generalization}, which measures the ability to predict other items rated by the same user, and \emph{strong generalization}, which measures the ability to predict items by novel users. 5,000 users are randomly selected for the weak generalization, and likewise 1,000 users are extracted for the strong generalization. Each experiment is then run three times and the averaged results are reported.  The performance metric is \emph{normalized mean absolute error} (NMAE) given as
\begin{equation*}
NMAE=\frac{\left(\sum_{i,j\in supp(\bX_0)}{\frac{|({\bX_0})_{ij}-\hat \bX_{ij}|}{|supp(\bX_0)|}}\right)}{(rt_{max}-rt_{min})},
\end{equation*}
where $rt_{max}$ and $rt_{min}$ are the the maximum and minimum ratings possible.

We followed the same setup and reported results using BARM in Table \ref{tab:movie} along with results from \cite{leger2010efficient} for comparison.  This includes the additional algorithms URP \cite{marlin2004collaborative}, Attitude \cite{marlin2003modeling}, MMMF \cite{rennie2005fast}, IPCF \cite{park2007applying}, E-MMMF \cite{decoste2006collaborative}, GPLVM \cite{lawrence2009non}, NBMC \cite{zhou2010nonparametric}, and IRLS/GM \cite{leger2010efficient,mohan2012iterative}.  From this table we observe that for the easier weak generalization problem BARM is a close second best, while for the more challenging strong generalization BARM is actually the best.  Of course it is also immediately apparent that all algorithms fall within a relatively narrow performance range of approximately five percentage points.  Consequently, we cannot unequivocally conclude that the attributes of BARM which make it suitable for optimally minimizing rank necessarily translate into a truly significant practical advantage on this collaborative filtering task.  But we would argue that the same holds for any matrix completion algorithm.

\section{Conclusion}

This paper explores a conceptually-simple, parameter-free algorithm called BARM for matrix rank minimization under affine constraints that is capable of successful recovery empirically observed to approach the theoretical limit over a broad class of experimental settings (including many not shown here) unlike existing algorithms, and long after any nuclear norm recovery guarantees break down.  Our strategy in this effort has been to adopt Bayesian machinery for inspiring a principled cost function; however, ultimate model justification is placed entirely in theoretical evaluation of desirable global and local minima properties, and in the empirical recovery performance that inevitably results from these properties.  Although in general non-convex algorithms are exponentially more challenging to analyze, in this regard we have at least attempted to contextualize BARM in the same manner as convex optimization-based approaches such as nuclear-norm minimization.

% All of this suggests that improved algorithms may be possible if we discard the Bayesian

%Perhaps one of the interesting aspects of the proposed BARM algorithm is that it arises from such simple Bayesian origins related to probabilistic PCA \cite{Tipping1999PPCA}.  We emphasize though that Bayesian inspiration can take uncountably many different forms and parameterizations, and it is not at all immediately obvious which instantiation might be optimal.  For example \cite{babacan2012sparse} adopts low-rank-favoring priors and factorizations and then applies related variational inference techniques, and yet performance is not nearly as high as BARM on canonical matrix completion tasks.
%

%While our model is ultimately based on a Gaussian marginal likelihood function, variations of which have been analyzed thoroughly in the context of sparse estimation \cite{wipf2011latent}, the affine rank minimization problem addressed here is considerably different.  Moreover, our main theoretical result, Theorem \ref{thm:local_min_theorem}, relies on a completely different underlying analysis technique; likewise for the symmeterization adaptation.  This mirrors well-established differences between convex $\ell_1$ and nuclear norm algorithms for compressive sensing and rank minimization respectively.

%\appendix

\section*{Appendix}
Here we provide brief proofs of Lemmas 1 and 2 as well as Theorem 1. We also address the augmented update rules that account for the revised, symmetrized cost function discussed in Section \ref{sec:symm}.
%\textcolor{red}{We also provide empirical comparisons with alternating algorithms given correct rank.}

%, followed by a brief discussion of the associated computational complexity.

\subsection{Proof of Lemmas 1 and 2:}

Regarding Lemma 1, this result mirrors related ideas from \cite{wipf2011latent} in the context of Bayesian compressive sensing.  Hence, while a more rigorous presentation is possible, here we describe the basic aspects of the adaptation.  At any candidate minimizer of (\ref{eq:BARM_cost}) in the limit $\lambda \rightarrow 0$, define $\bW$ such that  $\bA \bar{\bPsi} \bA^{\top} = \bW \bW^{\top}$.  To be a minimizer, global or local, it must be that $\bb \in \myspan[\bW]$. If this were not the case, then $\calL(\bPsi,\bnu)$ would diverge to infinity as $\lambda \rightarrow 0$ because $\bb^T \bSigma_b^{-1} \bb$ progresses to infinity at a faster rate than $\log|\bSigma_b|$ can compensate by approaching minus infinity.  Intuitively, in much the same way $\arg\min_z \frac{1}{z} + \log z = 1$, meaning the optimal $z$ must lie in the `span' of 1 else the overall objective will be driven to infinity.

%$\log|\bSigma_b|$  term is not a sufficiently strong counterbalance, in the same way that $\min_z \frac{1}{z} + \log z$ equals one (finite).

Consequently, the only way to minimize the cost in the limit as $\lambda \rightarrow 0$ is to consider low-rank solutions within the constraint set that $\bb \in \myspan[\bW]$, and it is equivalent to requiring that $\bb^T \bSigma_b^{-1} \bb \leq C$ for some constant $C$ independent of $\lambda$ (which ultimately corresponds with maintaining $\calA(\bX) = \bb$ in the limit as well).

In this setting, while $0 \leq \bb^T \bSigma_b^{-1} \bb \leq C$ is bounded, the second term in $\calL(\bPsi,\bnu)$ can be unbounded from below when  $\rank[\bPsi]$ is sufficiently small.  To see this note that
\begin{equation} \label{eq:log_det_term}
\log|\bSigma_b| = \sum_{i=1}^p \log \left(\sigma_i[\bA \bar{\bPsi} \bA^{\top} ] + \lambda    \right),
\end{equation}
where $\sigma_i[\cdot ]$ denotes the $i$-th singular value of a matrix.  While the maximum rank of $\bA \bar{\bPsi} \bA^{\top}$ is obviously $p$, if $r \triangleq \rank[\bPsi] < p/m$ and $\spark[\bA] = p+1$ (maximal spark) as stipulated in the lemma statement, then $\rank[\bA \bar{\bPsi} \bA^{\top}] = m r$ and (\ref{eq:log_det_term}) becomes
\begin{equation} \label{eq:log_det_term2}
\log|\bSigma_b| = \sum_{i=1}^{mr} \log \left(\sigma_i[\bA \bar{\bPsi} \bA^{\top} ] + \lambda  \right) + (p-mr) \log \lambda.
\end{equation}
Note that the spark assumption accomplishes two objectives in this context.  First, it guarantees that a high rank $\bPsi$ cannot masquerade as a low rank $\bPsi$ behind the nullspace of some collection of columns $\bA_i$.  Secondly, it ensures that after assuming $r < p/m$, then $\rank[\bA \bar{\bPsi} \bA^{\top}] = m r$.

Consequently, in the limit where $\lambda \rightarrow 0$ (with the limit being taken outside of the minimization), (\ref{eq:log_det_term}) effectively scales as $(p-mr) \log \lambda$, and hence the overall cost is minimized when $\bPsi$ has minimal rank.  This in turn ensures that the corresponding $\bX$ will also have minimal rank, completing the proof sketch for Lemma 1.

Finally, Lemma 2 follows directly from the structure of the $\calL(\bPsi,\bnu)$ cost function via simple reparameterizations.  \myendofproof

\subsection{Proof of Theorem 1:}

To begin we assume that $\bb_i \neq 0$, $\forall i$, where $\bb_i$ denotes the sub-vector of $\bb$ such that $\bb_i = \bA_i \bx_{:i}$.  If this were not the case we can always collapse $\bX$ by the corresponding column (which is indistinguishable from zero) and achieve an equivalent result. Given the assumptions of Theorem 1, the BARM cost function becomes
\begin{equation}
\mathcal{L}(\bPsi,\bnu) = \sum_{i=1}^m \bb_i^{\top} \left( \nu_i \bA_i  \bPsi \bA_i^{\top}  \right)^{-1} \bb_i + \log \left| \nu_i \bA_i  \bPsi \bA_i^{\top}  \right|.
\end{equation}
If there exists a feasible rank one solution to $\bb = \bA \myvec[\bX]$, then there also exists a set of $\bPsi_i' = \nu_i \bPsi$ such that $\bb_i \bb_i^{\top} = \bA_i \bPsi_i' \bA_i^{\top}$ for all $i$.  To see this, note that $\bb_i \bb_i^{\top} = \bA_i \bx_{:i} \bx_{:i}^{\top} \bA_i^{\top}$.  Because $\rank[\bX] = 1$, it also follows that $\bb_i \bb_i^{\top} = \alpha_i \bA_i \bX \bX^{\top} \bA_i^{\top}$, where $\alpha_i = \| \bx_{:i} \bx_{:i}^{\top} \|/\| \bX \bX^{\top} \|$.  Therefore $\bPsi_i' = \nu_i \bX \bX^{\top}$ achieves the desired result with $\nu_i = \alpha_i$.

Now suppose we have converged to any solution $\{\hat{\bPsi},\hat{\bnu}\}$ with $\rank[\bPsi] > 1$ and associated $\bar{\hat{\bPsi}} = \bI \otimes \hat{\bPsi}$.  Note that since $\bb_i \neq 0$, $\bnu_i > 0$ for all $i$, otherwise a local minimum is not possible (the cost function would be driven to positive infinity).

Define $\hat{\bSigma}_{b_i} = \hat{\nu}_i \bA_i  \hat{\bPsi} \bA_i^{\top}$.  Additionally we can assume that $\bb_i^{\top} \hat{\bSigma}^{-1}_{b_i}$ is finite, meaning that $\bb_i$ lies in the span of the singular vectors of $\hat{\bSigma}_{b_i}$.  (If this were not the case, the cost would be driven to infinity and we could not be at a minimizing solution anyway.)  If $\{\hat{\bPsi},\hat{\bnu}\}$ is a local minimum, then $\{\lambda_1 = 1, \lambda_2 = 0\}$  must be a local minimum of the revised cost function
\begin{equation}
\hspace*{-1.0cm} \mathcal{L}(\lambda_1,\lambda_2) = \sum_{i=1}^m \bb_i^{\top} \left( \lambda_1 \hat{\bSigma}_{b_i} + \lambda_2 \bb_i \bb_i^{\top}   \right)^{-1} \bb_i \nonumber
\end{equation}
\begin{equation}
\hspace*{0.5cm} + \log \left| \lambda_1\hat{\bSigma}_{b_i} + \lambda_2 \bb_i \bb_i^{\top}   \right|.
\end{equation}
This is because $\bb_i \bb_i^{\top}$ represents a valid set of basis vectors for updating the covariance per the construction above involving $\bPsi'_i$. First consider optimization over $\lambda_1$.  If $\lambda_1 = 1$ is a local minimum, then by taking gradients and equating to zero, we require that
\begin{equation}
\sum_{i=1}^m \bb_i^{\top} \hat{\bSigma}_{b_i}^{-1} \bb_i = \sum_{i=1}^m \rank[\hat{\bSigma}_{b_i}].
\end{equation}
Likewise, taking the gradient with respect to $\lambda_2$ we obtain
\begin{equation} \label{eq:beta_deriv}
\left. \frac{\partial \mathcal{L}(\lambda_1,\lambda_2)}{\partial \lambda_2} \right|_{\lambda_1 = 1, \lambda_2 = 0} = \sum_{i=1}^m \bb_i^{\top} \hat{\bSigma}_{b_i}^{-1} \bb_i - \sum_{i=1}^m  \left( \bb_i^{\top} \hat{\bSigma}_{b_i}^{-1} \bb_i \right)^2.
\end{equation}
The nullspace condition (a very mild assumption) ensures that $\sum_{i=1}^m \rank[\hat{\bSigma}_{b_i}] = k$ for some $k> m$ when $\rank[\bPsi] > 1$.  To see this, observe that to achieve $\sum_{i=1}^m \rank[\hat{\bSigma}_{b_i}] = m$ when $\rank[\bPsi] > 1$ requires that $\bPsi = \bu \bu^{\top} + \bW \bW^{\top}$ where $\bu$ is a vector and $\bW$ is a matrix (or vector) with columns in $\mbox{null}[\bA_i]$, $\forall i$.  If any such $\bW$ is not in this nullspace for some $i$, then given that $p_i > 1$, the associated $\bA_i \bPsi \bA_i^{\top}$ will have rank greater than one, and the overall rank sum will exceed $m$.

Consequently, (\ref{eq:beta_deriv}) will always be negative.  This is because if $\sum_{i=1}^m z_i = k$ for any set of non-negative variables $\{z_i\}$, the minimal value of $\sum_{i=1}^m z_i^2$ occurs when $z_i = k/m$, $\forall i$.  In our case, this implies that
\begin{equation}
\sum_{i=1}^m  \left( \bb_i^{\top} \hat{\bSigma}_{b_i}^{-1} \bb_i \right)^2 \geq \sum_{i=1}^m (k/m)^2 > k > m.
\end{equation}
Therefore we can add a small contribution of $\bb_i \bb_i^{\top}$ to each $\hat{\bSigma}_{b_i}$ and reduce the underlying cost function.  Hence we cannot have a local minimum, except when $\bPsi$ is equal to some $\bPsi^*$ with $\rank[\bPsi^*] = 1$.  Moreover, we may directly conclude that $\bx^* = \bar{\bPsi}^* \bA^{\top} \left( \bA \bar{\bPsi}^* \bA^{\top}\right)^{\dag} \bb$ is feasible and $\rank[\bX^*] = 1$ with $\bx^* = \myvec[\bX^*]$.

Regarding the last part of the theorem, we consider only $f$ that are concave non-decreasing functions (this is the only reasonable choice for shrinking singular values to zero, and the more general case naturally follows anyway with additional effort, but minimal enlightenment).  Without loss of generality we may also assume that $f(0) = 0$ and $f(1) = 1$; we can always apply an inconsequential translation and scaling such that these conditions hold.\footnote{The $\log$ function is a limiting case, but what follows holds nonetheless.}   Simple counter examples then demonstrate that $f(\epsilon)$ must be greater than some constant $C$ independent of $\epsilon$ for all $\epsilon$ sufficiently small.  To see this, note that we can always rescale elements of $\bA$ such that a solution with rank greater than one is preferred unless this condition holds.  However, such an $f$, which effectively must display infinite gradient at $f(0)$ to guarantee a global solution is always rank one, will then always display local minima for certain $\bA$.  This can easily be revealed through simple counter-examples. \myendofproof

\subsection{Symmetrization Update Rules}

%As discussed in the main paper, the symmetrized version of BARM assumes that $\bx = \myvec[\bX]$ with a covariance formed using  block-wise averaging of co-variances defined over both rows and columns, denoted $\bPsi_r$ and $\bPsi_c$ respectively. For easy repeatition of this work, we elaborate the corresponding updating rules as follows.

% Here we provide the update rules for the symmetrized version of BARM discussed in the main paper.

These iterative update rules follow from alternative upper bounds tailored to the symmetric version of BARM.  When both $\bPsi_r$ and $\bPsi_c$ are fixed, $\bx$ is updated via the posterior mean calculation
\begin{equation} \label{eq:posterior_mean_sym}
\begin{split}
& \hat{\bx} = \myvec[\hat{\bX}]  \\
& = \frac{1}{2}( \bar{\bPsi}_r+\bar{\bPsi}_c) \bA^{\top} \left[\lambda \bI + \bA \frac{1}{2}\left(\bar{\bPsi}_r +\bar{\bPsi}_c \right) \bA^{\top} \right]^{-1} \bb.
\end{split}
\end{equation}
where $\bar{\bPsi}_r =   \bPsi_r \otimes \bI$ and $\bar{\bPsi}_c = \bI \otimes \bPsi_c$.  Likewise we update $\nabla_{\Psi_r^{-1}}$ and $\nabla_{\Psi_c^{-1}}$ using
\begin{equation} \label{eq:posterior_cov_sym}
\nabla_{\Psi_r^{-1}} = \sum_{i=1}^m  \bPsi_r - \bPsi_r \bA_{ri}^{\top} \left(  \bA \bar{\bPsi}_r \bA^{\top} + \lambda \bI   \right)^{-1}  \bA_{ri} \bPsi_r,
\end{equation}
\begin{equation} \label{eq:posterior_cov2_sym}
\nabla_{\Psi_c^{-1}} = \sum_{i=1}^n  \bPsi_c - \bPsi_c \bA_{ci}^{\top} \left(  \bA \bar{\bPsi}_c \bA^{\top} + \lambda \bI   \right)^{-1}  \bA_{ci} \bPsi_c,
\end{equation}
where $\bA_{ri} \in \mathbb{R}^{p\times m}$ is defined such that $\bA = [\bA_{r1}^\top,\ldots,\bA_{rm}^{\top}]^\top$ and $\bA_{ci} \in \mathbb{R}^{p\times m}$ is defined such that $\bA = [\bA_{c1}, \ldots,\bA_{cn}]$.  Finally given these values, with $\bX$, $\nabla_{\Psi_r^{-1}}$ and $\nabla_{\Psi_c^{-1}}$ fixed, we can compute the optimal $\bPsi_r$ and $\bPsi_c$ in closed form by optimizing the relevant $\bPsi_r$- and $\bPsi_c$-dependent terms via
\begin{gather} \label{eq:psi_update_sym}
\bPsi_r^{opt}  = \frac{1}{n} \left[\hat{\bX}^{\top} \hat{\bX} + \nabla_{\Psi_r^{-1}} \right], \\
\bPsi_c^{opt}  = \frac{1}{m} \left[\hat{\bX} \hat{\bX}^{\top} + \nabla_{\Psi_c^{-1}} \right].
\end{gather}
In practice the simple initialization $\bPsi_r = \bI$ and $\bPsi_c = \bI$ is sufficient for obtaining good performance.

\ifCLASSOPTIONcaptionsoff
  \newpage
\fi



%\bibliographystyle{IEEEtran}
%\begin{thebibliography}{1}
%\end{thebibliography}

\bibliographystyle{IEEEtran}
\bibliography{refr}

\end{document}